\documentclass{article}

\usepackage[left=2.40cm, right=2.40cm, top=2.40cm, bottom=2.40cm]{geometry}

\usepackage[titletoc,title]{appendix}

\usepackage[numbers]{natbib}

\usepackage{hyperref}       
\usepackage{url}            
\usepackage{booktabs}       
\usepackage{amsfonts}       
\usepackage{nicefrac}       
\usepackage{microtype}      

\usepackage{amsmath}
\usepackage{amsthm}
\usepackage{amssymb}
\usepackage{graphicx}
\usepackage{caption}
\usepackage{subfig}
\usepackage{framed}
\usepackage{bm}             

\title{Dictionary Learning by Dynamical Neural Networks}

\DeclareMathOperator*{\argmin}{arg\,min}
\newcommand{\bydef}{\stackrel{\mathrm{def}}{=}}

\newtheorem{mythm}{Theorem}

\newtheorem*{myproof}{Proof}

\newcommand{\VS}{}

\newcommand{\tendsto}{\rightarrow}

\newcommand{\diag}{{\rm diag}}

\newcommand{\reals}{\mathbb{R}}

\newcommand{\suchthat}{\,\mid\,}

\newcommand{\Enormsq}[1]{\| {#1} \|_2^2}
\newcommand{\Fnormsq}[1]{\| {#1} \|_F^2}
\newcommand{\Onenorm}[1]{\| {#1} \|_1}
\newcommand{\Maxnorm}[1]{\| {#1} \|_\infty}

\newcommand{\oneminus}[1]{{#1}^c}
\newcommand{\perturbed}[1]{\hat{{#1}}}

\newcommand{\ssup}[2]{{#1}^{(#2)}}

\newcommand{\bb}[1]{\boldsymbol{#1}}
\newcommand{\vv}[1]{\mathbf{#1}}

\newcommand{\vva}{\vv{a}}

\newcommand{\vvd}{\vv{d}}
\newcommand{\vve}{\vv{e}}

\newcommand{\vvs}{\vv{s}}

\newcommand{\vvx}{\vv{x}}

\newcommand{\vvzero}{\vv{0}}

\newcommand{\cc}[1]{{\cal #1}}
\newcommand{\ccA}{\cc{A}}

\newcommand{\ccI}{\cc{I}}

\newcommand{\curr}{\mu}
\newcommand{\bias}{\beta}
\newcommand{\poten}{\rho}
\newcommand{\thres}{\theta}
\newcommand{\spike}{\delta}
\newcommand{\spiketrain}{\sigma}
\newcommand{\avgcurr}{u}
\newcommand{\spikerate}{a}

\newcommand{\currvec}{\boldsymbol{\mu}}
\newcommand{\biasvec}{\boldsymbol{\beta}}

\newcommand{\thresvec}{\boldsymbol{\theta}}
\newcommand{\Thres}{\Theta}
\newcommand{\spiketrainvec}{\boldsymbol{\sigma}}
\newcommand{\avgcurrvec}{\vv{u}}
\newcommand{\spikeratevec}{\vv{a}}

\newcommand{\spt}[2]{t_{#1,#2}}

\usepackage{algorithm}
\usepackage{algorithmic}

\newtheorem{appxthm}{Theorem}[section]

\author{
	Tsung-Han Lin and Ping Tak Peter Tang \\ \\
    Intel Corporation \\
	\{tsung-han.lin, peter.tang\}@intel.com     
}

\date{}

\begin{document}

\maketitle

\begin{abstract}
A \textit{dynamical neural network} 
consists of a set of interconnected neurons that interact over time continuously.
It can exhibit computational properties
in the sense that the dynamical system's evolution and/or limit points in the
associated state space can correspond to numerical solutions to certain 
mathematical optimization or learning problems.
Such a computational system is particularly attractive in that it can be mapped to 
a massively parallel computer architecture for power and throughput efficiency,
especially if each neuron can rely solely on local information (i.e., local memory).
Deriving gradients from the dynamical network's various states 
while conforming to this last constraint, however, is challenging.
We show that by combining ideas of \textit{top-down feedback} and \textit{contrastive learning}, a dynamical network for solving the $\ell_1$-minimizing dictionary learning problem can be constructed, and the true gradients for learning are provably computable by individual neurons.
Using spiking neurons to construct our dynamical network, we present a learning
process, its rigorous mathematical analysis, and numerical results on several
dictionary learning problems. 


\end{abstract}

\section{Introduction}

A network of simple neural units can form a physical system that exhibits computational properties. 
Notable examples include Hopfield network \cite{hopfield82} and Boltzmann machine \cite{ackley1985learning}.
Such systems have global states that evolve over time through the interactions among local neural units.
Typically, one is interested in a system whose motion converges towards locally stable limit points, 
with the limit points representing the computational objective of interest. For example, 
a Hopfield network's limit points correspond to stored memory information and that of a
Boltzmann machine, a data representation.
These computational systems are particularly interesting from a hardware implementation standpoint. 
A subset of the neurons can be mapped to one processing element in a massively parallel architecture~\cite{davies2018loihi,merolla2014million}.
By allocating private local memory to each processing element, 
the so-called von Neumann memory bottleneck in modern computers can be eliminated~\cite{kung1982systolic}.

We are interested in using such systems to solve the 
$\ell_1$-minimizing sparse coding and dictionary learning problem, 
which has fundamental importance in many areas, e.g., see \cite{mairal2014sparse}.
It is well-known that even just the sparse coding problem, with a prescribed dictionary,
is non-trivial to solve, mainly due to the
non-smooth objective involving an $\ell_1$-norm~\cite{efron2004least,beck2009fast}.
It is therefore remarkable that a dynamical network 
known as the LCA network \cite{rozell2008sparse} can be carefully constructed 
so that its limit points are identical to the solution of the sparse coding problem.
Use of a dynamical network thus provides an alternative and potentially more power efficient
method for sparse coding to standard numerical optimization techniques.
Nevertheless, while extending numerical optimization algorithms 
to also learning the underlying dictionary is somewhat straightforward, 
there is very little understanding in using dynamical networks to learn a dictionary with provable guarantees.

In this work, we devise a new network topology and learning rules that enable dictionary learning.
On a high level, our learning strategy is similar to the contrastive learning procedure 
developed in training Boltzmann machines, which also gathers much recent interest in deriving implementations of
backpropagation under neural network locality constraints~\cite{ackley1985learning,movellan1991contrastive,o1996biologically,xie2003equivalence,scellier2017equilibrium,whittington2017approximation}.
During training, the network is run in two different configurations -- a ``normal'' one
and a ``perturbed'' one.\footnote{In Botlzmann machine, the two configurations are called 
the \textit{free-running phase} and the \textit{clamped phase}.}
The networks' limit points under these two configurations will
be identical if the weights to be trained are already optimal, but different 
otherwise.
The learning process is a scheme to so adjust the weights to minimize the difference in
the limit points. In Boltzmann machine, the weight adjustment can be 
formulated as minimizing a KL divergence objective function.

For dictionary learning, we adopt a neuron model whose activation function corresponds to the unbounded ReLU function
rather than the bounded sigmoid-like function in Hopfield networks or Boltzmann machines, and
a special network topology where connection weights have dependency. 
Interestingly, the learning processes are still similar: We also rely on running our network in two configurations. The difference in states 
after a long-enough evolution, called limiting states in short,
is shown to hold the gradient information of a dictionary learning objective function which
the network minimizes, as well as the gradient information for the network to maintain weight dependency.



\begin{figure}[t]
	\center
	\setlength\tabcolsep{1.5pt}
	\begin{tabular}{cc}
		\begin{tabular}{c}
		\includegraphics[scale=0.35]{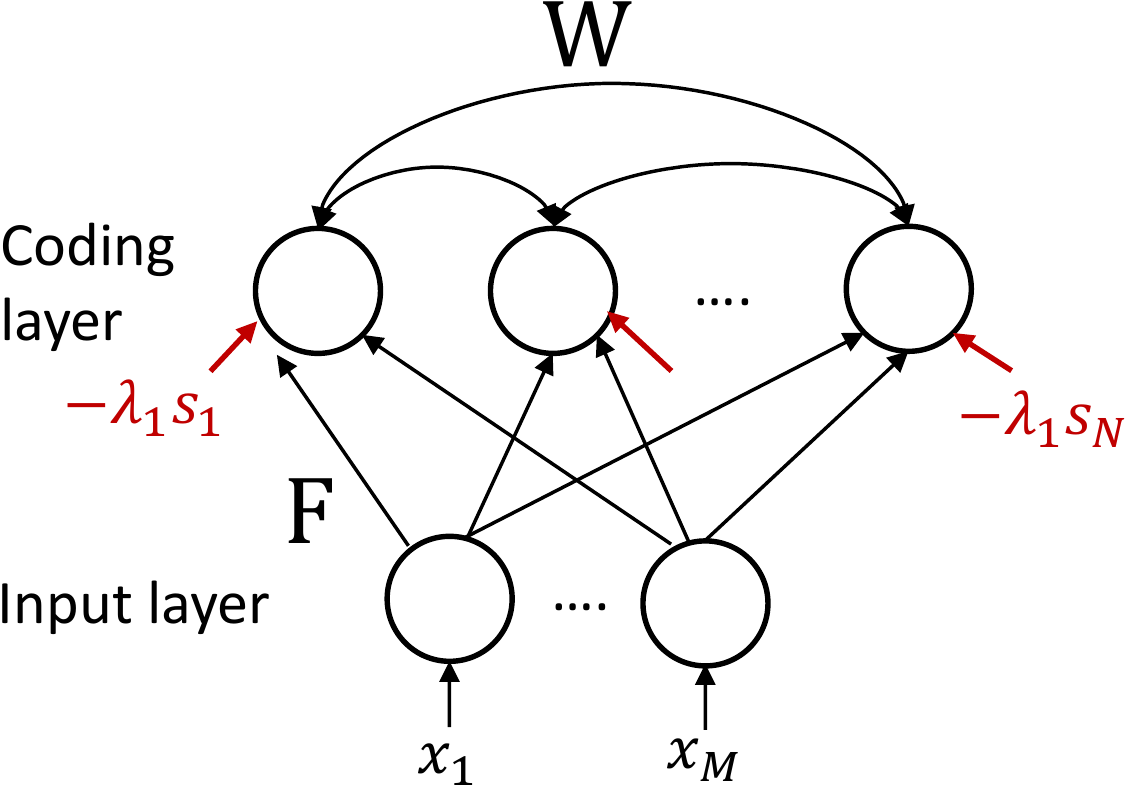} \\ (a) 
		\end{tabular}	
		&
		\begin{tabular}{c}
			\includegraphics[scale=0.35]{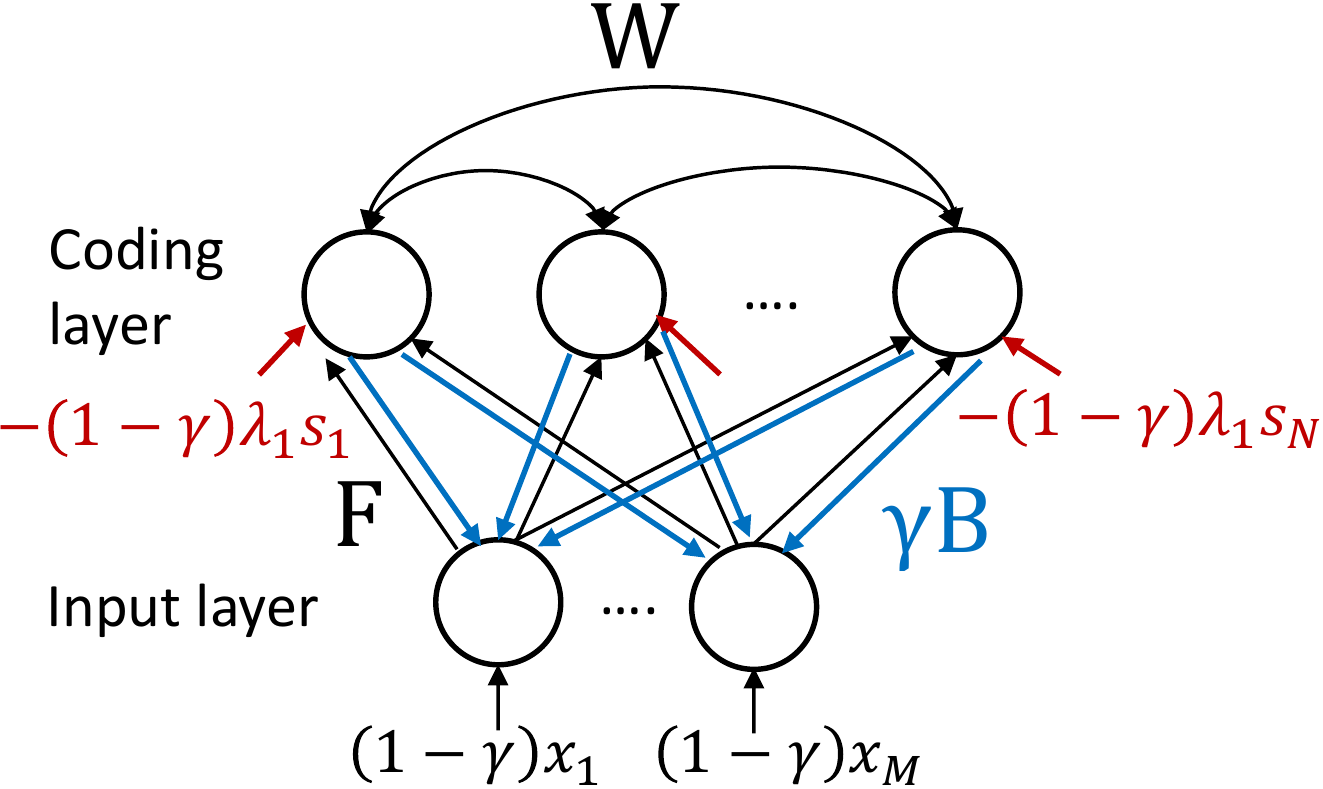} \\ (b) 
		\end{tabular}
	\end{tabular}	
	\caption{The network topologies discussed in this work. (a) is known as the LCA network that can perform sparse coding. We propose the network in (b) for dictionary learning.
	}
	\label{fig:networks}
\end{figure}

\subsection{Related Work}
\label{sec:related_work}
Dictionary learning is thought to be related to the formation of receptive fields in 
visual cortex~\cite{olshausen1996emergence}.
The typical architecture studied 
is a feedforward-only, two-layer neural network with inhibitory lateral connections among the second layer neurons~\cite{foldiak1990forming,zylberberg2011sparse,brito2016nonlinear,hu2014hebbian,seung2017correlation,vertechi2014unsupervised,brendel2017learning}, as shown in Figure~\ref{fig:networks}(a). 
The lateral connections allow the coding neurons to compete among themselves and hence induce sparseness in neural activities, giving dynamics more complex than conventional deep neural networks which do not have intra-layer connections.\footnote{This should not be confused with the conventional recurrent neural networks. Although RNNs also have intra-layer connections, these connections are still uni-directional over a sequence of input.}
In~\cite{rozell2008sparse}, it is shown that the coding neuron activations can correspond to a sparse coding solution if the connection weights are set according to a global dictionary $D$ as $F=D^T$, $W=-D^TD + I$.\footnote{The exact formulation depends on the neuron model. In the spiking neuron formulation, we in fact have $W-\Thres=-D^TD$ where $\Thres$ is the firing thresholds. 
See Section~\ref{sec:sparse_coding} for more details.}
To enable learning in this network (that is, each neuron locally adjusts their connection weights to adapt the dictionary; see Section~\ref{sec:parallel} for the definition of weight locality), one must address the following two questions:                                  
\begin{itemize}
	\setlength{\itemsep}{0pt}	
	\item How does individual neuron compute the gradient for learning locally?
	\item How do the neurons collectively maintain the global weight consistency between $F$ and $W$?
\end{itemize}

The first line of work~\cite{foldiak1990forming,zylberberg2011sparse,brito2016nonlinear} adopts the Hebbian/anti-Hebbian heuristics for learning the feedforward and lateral weights, respectively, and empirically demonstrated that such learning yielded Gabor-like receptive 
fields if trained with natural images.
However, unlike the network in~\cite{rozell2008sparse}, this learning heuristic does not correspond to a rigorous learning objective, and hence cannot address any of the two above questions.
Recently, this learning strategy 
is linked to  minimizing a similarity matching objective function between input and output correlations~\cite{hu2014hebbian}.
This formulation is somewhat different from the common autoencoder-style dictionary 
learning formulation discussed in this work. 

Another line of work~\cite{vertechi2014unsupervised,brendel2017learning} notes the importance of \textit{balance} between excitation and inhibition among the coding neurons, and proposes that the learning target of lateral connections should be to maintain such balance; that is, the inhibitory lateral weights should grow according to the feedforward excitations.
This idea provides a solution to ensure weight consistency between $F$ and $W$.
Nevertheless, similar to the first line of work, both~\cite{vertechi2014unsupervised,brendel2017learning} resort to pure Hebbian rule when 
learning the feedforward weights $F$ (or equivalently, learning the dictionary), which does not necessarily follow a descending direction that minimizes the dictionary learning objective function.

\subsection{Contributions}

The major advance in this work is to recognize the inadequacy of the customary 
feedforward-only architecture, and to introduce \textit{top-down feedback} connections shown in Figure~\ref{fig:networks}(b).  
As will later be shown, this network structure allows the true learning gradients to be provably computable from the resulting network dynamics.
Further, the existence of feedback allows us to devise a separate mechanism that acts as an inner loop during learning to continuously ensure weight consistency among all connections. 
Combining these two, we can successfully address both the above questions and the dictionary learning problem.

We will focus our discussion on a network that uses \textit{spiking neurons} as the basic 
units that are suited for digital circuit implementations with high computational efficiency.
Note that this does not result in a loss of generality.
The principles of LCA network can be applied to both continuous-valued and spiking neurons~\cite{shapero2014optimal,tang2016sparse},
and similarly the results established in this paper can be easily applied to construct a network of 
continuous-valued neurons for dictionary learning.


\section{Background}


\subsection{Integrate-and-Fire Spiking Neuron Model and Network Dynamics}
\label{sec:neuron_model}

An integrate-and-fire neuron has two internal state variables that govern its dynamics: 
the \textit{current} $\curr(t)$
and the \textit{potential} $\poten(t)$.
The key output of a neuron is a time sequence of spikes -- spike train -- that it produces.
A neuron's spike train is generated by its potential $\poten(t)$; $\poten(t)$ is
in turn driven by the current $\curr(t)$, which is in turn driven by a 
\textit{constant bias} $\bias$ (bias in short) and the spike trains of other neurons
to which it is connected. Specifically, each neuron has a configured firing threshold
$\thres > 0$. When $\poten(t)$ reaches $\thres$, say at time $t_k$, a spike 
given by the Dirac delta function $\spike(t-t_k)$ is generated and $\poten(t)$ is 
reset to 0: $\poten(t_k^+) = 0$. For
$t > t_k$ and before $\poten(t)$ reaches $\thres$ again, $\poten(t) = \int_{t_k}^t \curr(s)\,ds$.

In a system of $N$ neurons $n_i$, $i=1,2,\ldots,N$, 
let
$\spiketrain_j(t) = \sum_k \spike(t-t_{j,k})$ 
denote the spike train of neuron $n_j$. 
The current $\curr_i(t)$ of $n_i$ is given in terms of
its bias $\bias_i$ and the spike trains $\{\spiketrain_j(t)\}$: 
\begin{equation}
\label{eqn:soma_current}
\textstyle \curr_i(t) = \bias_i + \sum_{j\neq i} W_{ij}\,(\alpha \ast \spiketrain_j)(t),
\end{equation}
where $\alpha(t) = \frac{1}{\tau}e^{-t/\tau}$ for $t\ge 0$, $\alpha(t) = 0$ for $t < 0$ and
$\ast$ is the convolution operator. Neuron $n_j$ inhibits (excites) $n_i$ if $W_{ij} < 0$
($W_{ij} > 0$). If $W_{ij} = 0$, neurons $n_i$ and $n_j$ are not connected.
For simplicity, we consider
only $\tau=1$ throughout the paper.
Equation~\ref{eqn:soma_current} yields the dynamics
\begin{equation}
\label{eqn:current_dot}
\textstyle
\dot{\currvec}(t) = \biasvec - \currvec(t) + W\cdot\spiketrainvec(t),
\end{equation}
where the vectors $\currvec(t)$ and $\spiketrainvec(t)$ denote the $N$ currents
and spike trains.

The network dynamics can studied via the filtered quantities of average current and spike rate:
\begin{equation}
\label{eqn:avgcur_spikerate_def}
\avgcurrvec(t) \bydef \frac{1}{t}\int_0^t \currvec(s)\,ds, \qquad
\spikeratevec(t) \bydef \frac{1}{t}\int_0^t \spiketrainvec(s)\,ds.
\end{equation}
In terms of $\avgcurrvec(t)$ and $\spikeratevec(t)$, Equation~\ref{eqn:current_dot} becomes
\begin{equation}
\label{eqn:dynamics_of_averages}
\dot{\avgcurrvec}(t) = 
\biasvec - \avgcurrvec(t) + W\,\spikeratevec(t) + (\currvec(0)-\avgcurrvec(t))/t 
\end{equation}
The trajectory $(\avgcurrvec(t),\spikeratevec(t))$ has interesting properties. In particular,
Theorem~\ref{thm:limits_of_slacks} below
(cf. \cite{tang2016sparse}) shows that any limit point
$(\avgcurrvec^*,\spikeratevec^*)$ satisfies 
$\avgcurrvec^*-\Thres\spikeratevec^* \le \vvzero$, $\spikeratevec^* \ge \vvzero$
and $(\avgcurrvec^*-\Thres\spikeratevec^*)\odot \spikeratevec^* = \vvzero$ where
$\odot$ is elementwise product. These properties are crucial to Section~\ref{sec:learning}.
\begin{mythm}
\label{thm:limits_of_slacks}
Let $\Thres = \diag(\thresvec)$, $\thresvec=[\theta_1,\theta_2,\ldots,\theta_N]$, then
\begin{equation}
\label{eqn:limits_of_slacks}
\avgcurrvec(t) - \Thres\spikeratevec(t) = \biasvec + (W-\Thres)\cdot\spikeratevec(t) + \bb{\Delta}(t)
\end{equation}
where $\max(\avgcurrvec(t),\vvzero)-\Thres\spikeratevec(t) \rightarrow \vvzero$
and $\bb{\Delta}(t)\rightarrow\vvzero$.
\end{mythm}
As with all other theorems, Theorem~\ref{thm:limits_of_slacks} is given in a conceptual form
where the corresponding rigorous ``$\epsilon$-$\delta$'' versions are detailed in the Appendix.

\subsection{Parallel Model of Dynamical Neural Networks}
\label{sec:parallel}
We view the dynamical network as a computational model where each neuron evolves in parallel and asynchronously.
One-sided communication in the form of a one-bit signal from Neuron $n_j$
to Neuron $n_i$ occurs only if the two are connected and only when the former
spikes. 
The network therefore can be mapped to a massively parallel architecture, such as~\cite{davies2018loihi}, where the connection weights are stored distributively in each processing element's (PE) local memory. 
In the most general case, we assume the architecture has the same number of PEs and neurons; each PE hosts one neuron and stores the weights connected towards this neuron, that is, each PE stores one row of the $W$ matrix in Equation \ref{eqn:current_dot}.
With proper interconnects among PEs to deliver spike messages, the dynamical network 
can be realized to compute sparse coding solutions.  

This architectural model imposes a critical weight locality constraint on learning algorithms for dynamical networks: The connection weights must be adjusted with rules that rely only on locally available information
such as connection weights,
a neuron's internal states, 
and the rate of spikes it receives. 
The goal of this paper is to enable dictionary learning under this locality constraint. 



\section{Dictionary Learning}
\label{sec:learning}
In dictionary learning, we are given $P$ images $\ssup{\vvx}{p}\in\reals_{\ge0}^M$, 
$p=1,2,\ldots,P$.  
The goal is to find a dictionary consisting of a prescribed number of $N$ atoms,
$D = [\vvd_1,\vvd_2,\ldots,\vvd_N]$, $D \in \reals^{M\times N}$ such that each of
the $P$ images can be sparsely coded in $D$. We focus here on non-negative
dictionary and formulate our 
minimization problem as
\begin{equation}
\label{eqn:dictionary_learning}
\argmin_{\ssup{\vva}{p} \ge \vvzero,
D \ge \vvzero}
\sum_{p=1}^P l(D,\ssup{\vvx}{p},\ssup{\vva}{p}), \textrm{\phantom{k}}
l(D,\vvx,\vva) = \frac{1}{2}\Enormsq{\vvx - D\vva} + \lambda_1 \Onenorm{S\vva}
+ \frac{\lambda_2}{2} \Fnormsq{D},
\end{equation}
$S$ being a positive diagonal scaling matrix.

Computational methods such as stochastic online training~\cite{aharon2008sparse}
is known to be effective
for dictionary learning. With this method, one iterates on the following two steps,
starting with a random dictionary.
\begin{enumerate}
\item Pick a random image $\vv{x} \leftarrow \ssup{\vv{x}}{p}$ and obtain
sparse code $\vv{a}$ for the current dictionary $D$ and image $\vv{x}$,
that is, solve Equation~(\ref{eqn:dictionary_learning}) with $D$ fixed.
\item Use gradient descent to update $D$ with a learning rate $\eta$. The 
gradient $\nabla_D$ with respect to $D$ is in a simple form and the update
of $D$ is 
\begin{equation}
\label{eqn:update_D}
D^{({\rm new})} \leftarrow
D - \eta \left((D\vv{a}-\vv{x})\vv{a}^T + \lambda_2 D \right).
\end{equation}
\end{enumerate}
Implementing these steps with a dynamical network is challenging. First,
previous works have only shown that Step 1 can be solved when the configuration
uses the dictionary $D$ in the feedforward connection weights and $D^T D$ as the lateral
connection weights (\cite{shapero2014optimal}, c.f. Figure~\ref{fig:networks}(a) and below). 
For dictionary learning, both sets of
weights evolve without maintaining this exact relationship, casting doubt if
Step 1 can be solved at all. Second, the network in Figure~\ref{fig:networks}(a) only has
$F=D^T$, rendering the needed term $D\spikeratevec$ uncomputable using information
local to each neuron. 
Note that in general, gradients to minimize certain objective functions in a neural network 
can be mathematically derived, but often times they cannot be computed locally, e.g., standard backpropagation
and general gradient calculations for spiking networks~\cite{huh2017gradient}.
We now show that our design depicted in Figure~\ref{fig:networks}(b) 
can indeed implement Steps 1 and 2 and solve dictionary learning.

\subsection{Sparse Coding -- Getting $\spikeratevec$}
\label{sec:sparse_coding}
Non-negative sparse coding (Equation~\ref{eqn:dictionary_learning} with $D$ fixed)
is a constrained optimization problem. The standard approach (cf. \cite{BoydVandenberghe04})
is to augment $l(D,\vv{x},\vv{a})$ with non-negative slack variables, with which the optimal solutions
are characterized by the KKT conditions. Consider now Figure~\ref{fig:networks}(b)
that has explicit feedback weights $B$ whose strength is controlled by a parameter $\gamma$.
Equation~\ref{eqn:limits_of_slacks}, reflecting the structure of the coding and input neurons,
takes the form:
\begin{equation}
\label{eqn:block_eqn}
\begin{aligned}
\left[
\begin{matrix}
\vv{e}_\gamma(t) \\
\vv{f}_\gamma(t)
\end{matrix}
\right]
 & \bydef
\left[
\begin{matrix}
\vv{u}_\gamma(t)-\Thres\vv{a}_\gamma(t) \\
\vv{v}_\gamma(t)-\vv{b}_\gamma(t)
\end{matrix}
\right]  =
\left[
\begin{matrix}
-(1-\gamma)\lambda_1\vv{s} \\
(1-\gamma)\vv{x}
\end{matrix}
\right]
+
\left[
\begin{matrix}
-H &  F \\
\gamma B   & -I 
\end{matrix}
\right]
\,
\left[
\begin{matrix}
\vv{a}_\gamma(t)\\
\vv{b}_\gamma(t)
\end{matrix}
\right] 
+
\bb{\Delta}(t)
\end{aligned}
\end{equation}
$(\vv{u}(t),\vv{v}(t))$ and
$(\vv{a}(t),\vv{b}(t))$ denote the average currents and
spike rates for the coding and input neurons, respectively, and
$H \bydef W + \Thres$. 
When $\gamma=0$, $F^T=B=D$, $H=FB=D^T D$ and at a limit point
$(\vv{e}_0^*,\vv{a}_0^*)$, the network is equivalent to Figure~\ref{fig:networks}(a).
Equation~\ref{eqn:block_eqn} is simplified and reduces to
$\vv{e}_0^* = -\lambda_1\vv{s} - D^T\,D \vv{a}_0^* + D^T\vv{x}$ and that
$\vv{e}_0^* \le \vvzero$, $\vv{a}_0^* \ge \vvzero$ and
$\vv{e}_0^*\odot\vv{a}_0^* = \vvzero$.
This shows that $\vv{a}_0^*$ and $-\vv{e}_0^*$ are the optimal primal and
slack variables that satisfy the KKT conditions. In particular
$\vv{a}_0^*$ is the optimal sparse code. 

We extend this previously established result~\cite{tang2016sparse} in several aspects: 
(1) $\gamma$ can be set to any values in $[0,1)$; all $\vv{a}_\gamma^*$ are the optimal sparse code,
(2) $H$ needs not be $FB$ exactly; $\|H-FB\|$ being small suffices,
and
(3) as long as $t$ is large enough, $\vv{a}_\gamma(t)$ solves an approximate
sparse coding problem. These are summarized as follows (where the rigorous
form is presented in the Appendix).
\begin{mythm}
\label{thm:perturbation}
Let $F^T=B=D$, $\gamma\in[0,1)$ and $\|H-FB\|$ be small. Then
for $t$ large enough, $\spikeratevec_\gamma(t)$ is close
to an exact solution $\tilde{\spikeratevec}$ to 
Equation~\ref{eqn:dictionary_learning} ($D$ fixed) with
$S$ replaced by $\tilde{S}$ where $\|S-\tilde{S}\|$ is small.
\end{mythm}
The significant implication is that despite slight discrepancies between
$H$ and $FB$, the average spike rate $\spikeratevec_\gamma(t)$ at $t$
large enough is a practical solution to Step 1 of the stochastic learning
procedure.

\subsection{Dictionary Adjustment -- Updating $F,B$ and $H$}
\label{sec:dictionary_adjustment}
To obtain the learning gradients, we run the network for a long enough time to sparse code twice: 
at $\gamma = 0$ and $\gamma = \kappa > 0$, obtaining
$\tilde{\vv{e}}_0$, $\tilde{\vv{e}}_\kappa$,
$\tilde{\vv{a}}_0$, $\tilde{\vv{a}}_\kappa$ and 
$\tilde{\vv{b}}_0$, $\tilde{\vv{b}}_\kappa$  at those two configurations. 
We use tilde to denote the obtained states and loosely call them as limiting states.
Denote $1-\kappa$ by $\oneminus{\kappa}$. 
\begin{mythm}
\label{thm:approx_for_DA}
The limiting states satisfy 
\begin{align}
\kappa(B \tilde{\vv{a}}_\kappa - \vv{x}) & \approx \vv{g}_D, \phantom{kkk}
\vv{g}_D \bydef \tilde{\vv{b}}_\kappa - \tilde{\vv{b}}_0 \label{eqn:limiting_states_b} \\
\kappa(H-FB)\tilde{\vv{a}}_\kappa & \approx \vv{g}_H, \phantom{kkk}
\vv{g}_H \bydef \oneminus{\kappa}H(\tilde{\vv{a}}_0 - \tilde{\vv{a}}_\kappa) +
(\oneminus{\kappa}\tilde{\vv{e}}_0 - \tilde{\vv{e}}_\kappa) 
\label{eqn:limiting_states_e}
\end{align}
\end{mythm}
We now show
Theorem~\ref{thm:approx_for_DA} 
lays the foundation for computing all the necessary gradients that we need.
Equation~\ref{eqn:limiting_states_b} shows that (recall $B=D$)
\[
D\tilde{\vv{a}}_\kappa - \vv{x}
\approx 
\kappa^{-1} \vv{g}_D.
\]
In other words, the spike rate differences at the input layer reflect 
the reconstruction error of the sparse code we just computed.
Following Equation~\ref{eqn:update_D}, this implies that the update to each weight can be approximated from the 
spike rates of the two neurons that it connects, while the two spike rates 
surely are locally available to the destination neuron that stores the weight.
Specifically, each coding neuron has a row of the matrix $F = D^T$; each input neuron has a row of the
matrix $B = D$. These neurons each updates its row of matrix via
\begin{equation}
\label{eqn:update_FB}
\begin{aligned}
F_{ij}^{({\rm new})} 
    & \gets F_{ij} - \eta_D\left( \kappa^{-1} (\tilde{\vv{a}}_\kappa)_i 
         \, 
   (\vv{g}_D)_j
 + \lambda_2 F_{ij} \right) \\
B_{ij}^{({\rm new})} 
    & \gets B_{ij} - \eta_D\left( \kappa^{-1} (\tilde{\vv{a}}_\kappa)_j 
         \, 
   (\vv{g}_D)_i
 + \lambda_2 B_{ij} \right)
\end{aligned}
\end{equation}
Note that $F^T=B=D$ is maintained.

Ideally, at this point the $W$ and $\Thres$ stored distributively in the coding neurons 
will be updated to $H^{({\rm new})}$
where
$H^{({\rm new})} = F^{({\rm new})} B^{({\rm new})}$. 
Unfortunately,
each coding neuron only possesses one row of the matrix $F^{({\rm new})}$
and does not have access to any values of the matrix $B^{({\rm new})}$. 
To maintain $H$ to be close to $D^T D$ throughout the learning process, we do the following.
First we aim to modify $H$ to be closer to $FB$ (not 
$F^{({\rm new})} B^{({\rm new})}$) by reducing the cost function
$\phi(H) = \frac{1}{2} \Enormsq{ (H-FB)\tilde{\vv{a}}_\kappa }$. The
gradient of this cost function is
$
\nabla_H \phi = (H-FB) \tilde{\vv{a}}_\kappa
\tilde{\vv{a}}^T_\kappa
$
which is computable as follows. Equation~\ref{eqn:limiting_states_e} shows that
\[
\nabla_H \phi \approx G \bydef \kappa^{-1} \vv{g}_H \tilde{\vv{a}}^T_\kappa
%
\]
Using this approximation, coding neuron $n_{C,i}$ has the information to compute
the $i$-th row of $G$.
We modify $H$ by $-\eta_H G$ where $\eta_H$ is some learning rate. This
modification can be thought of as a catch-up correction because
$F$ and $B$ correspond to the updated values from a previous iteration. Because
the magnitude of that update is of the order of $\eta_D$, we have 
$\|H-FB\|\approx\eta_D$ and $\|G\| \approx \eta_D$.
Thus $\eta_H$ should be bigger than $\eta_D$ lest
$\|\eta_H G\| \approx \eta_H \eta_D$ be too small to be an effective 
correction.
In practice, $\eta_H \approx 15 \eta_D$ works very well.

In addition to this catch-up correction, we also 
make correction of $H$ due to the update of $-\eta_D \lambda_2 F$
and $-\eta_D \lambda_2 B$ to $F$ and $B$. These updates
lead to a change of $-2\eta_D FB + O(\eta_D^2)$. Consequently, after
Equation~\ref{eqn:update_FB}, we update $H$ by
\begin{equation}
\label{eqn:update_H}
H_{ij}^{({\rm new})} 
     \gets H_{ij} - \eta_H 
\kappa^{-1}(\vv{g}_H)_i (\vv{a}_\kappa)_j
- 2\eta_D \lambda_2 H_{ij}.
\end{equation}
Note that the update to $H$ involves update to the weights $W$ as well as the thresholds $\Thres$ (recall that $H \bydef W + \Thres$).
Combining the above, we summarize the full dictionary learning algorithm below.

\begin{algorithm}
\begin{algorithmic}
\STATE \emph{Initialization:} Pick a random dictionary 
$D \ge \vvzero$ with atoms of
unit Euclidean norm. Configure $F\leftarrow D^T$, $B \leftarrow D$,
$\vv{s}\leftarrow [1,1,\ldots,1]^T$, and $H \leftarrow FB$.
\REPEAT
  \STATE 1. \emph{Online input:}
         Pick a random image $\vv{x}$ from $\{\ssup{\vv{x}}{p}\}$
  \STATE 2. \emph{Sparse coding:} 
            Run the network at $\gamma \leftarrow 0$ and at
            $\gamma \leftarrow \kappa>0$.
  \STATE 3. \emph{Dictionary update:} Compute 
the vectors $\vv{g}_D$ and $\vv{g}_H$ distributively according to
Equations~\ref{eqn:limiting_states_b} and
\ref{eqn:limiting_states_e}.
   Update $F$, $B$ and $H$ according to Equations~\ref{eqn:update_FB}
   and~\ref{eqn:update_H}. Project the weights to non-negative quadrant.
  \STATE 4. \emph{Scaling update:}
            Set the scaling vector $\vv{s}$ to
         ${\rm diag}(H)$. This scaling helps maintain each atom
         of the dictionary to be of similar norms.
\UNTIL{dictionary is deemed satisfactory}
\end{algorithmic}
\caption{Dictionary Learning}
\label{algo:dictionary_learning}
\end{algorithm}


\subsection{Discussions}
\label{sec:discussions}

\textbf{Dictionary norm regularization.}
In dictionary learning, typically one needs to control the norms of atoms to prevent them from growing arbitrarily large. 
The most common approach is to constrain the atoms to 
be exactly (or at most) of unit norms,
achieved by re-normalizing each atom after a dictionary update.
This method however cannot be directly adopted in our distributed setting.
Each input neuron only has a row of the matrix $B$ but not a column of $B$ --
an atom -- so as to re-normalize.

We chose instead to regularize the Frobenius norm of the dictionaries, translating to a simple decay term in the learning rules. 
This regularization alone may result in learning degenerate zero-norm atoms
because sparse coding tends to favor larger-norm atoms to be actively updated, leaving
smaller-norm ones subject solely to continual weight decays.
By choosing a scaling factor $\vv{s}$ set to $\diag(H)$, sparse coding favors
smaller-norm atoms to be active and effectively mitigates the problem of
degeneracy.

\textbf{Boundedness of network activities.}
Our proposed network is a feedback nonlinear system, and one may wonder whether the network activities will remain bounded. 
While we cannot yet rigorously guarantee boundedness and stability under some a priori conditions, 
currents and spike rates remain bounded throughout learning for all our experiments.
One observation is that the feedback excitation amounts to $\gamma FB\vv{a}_\gamma(t)$ and the inhibition is $H\vv{a}_\gamma(t)$. 
Therefore when $H= FB$ and $\gamma < 1$, the feedback excitation is nullified, keeping the network from growing out of bound.

\textbf{Network execution in practice.}
Theoretically, an accurate spike rate can only be measured at a very large $T$
as precision increases at a rate of $O(1/t)$.
In practice, we observed that a small $T$ suffices for dictionary learning purpose. 
Stochastic gradient descent is known to be very robust against noise and thus can tolerate the low-precision spike rates as well as the approximate sparse codes due to the imperfect $H \approx FB$. 
For faster network convergence, the second network $\gamma=\kappa$ is ran right after the first network $\gamma=0$ with all neuron states preserved. 

\textbf{Weight symmetry.} 
The sparse code and dictionary gradient are computed using the feedforward and feedback weights respectively. 
Therefore a symmetry between those weights is the most effective for credit assignment.
We have assumed such symmetry is initialized and the learning rules can subsequently maintain the symmetry.
One interesting observation is that even if the weights are asymmetric, our learning rules still will symmetrize them. 
Let $E_{ij}^{(p)} = F_{ji}^{(p)} - B_{ij}^{(p)}$ be the weight difference at 
the $p$-th iteration. It is straightforward to show
$E_{ij}^{(p)} = \alpha^{p-1} E_{ij}^{(1)}$,
$\alpha = 1 - \eta_D \lambda_2$.
Hence $E_{ij}^{(p)} \tendsto 0$ as $p$ gets bigger.
In training deep neural networks, symmetric feedforward and feedback weights are important for similar reasons. 
The lack of local mechanisms for the symmetry to emerge makes backpropagation biologically implausible and hardware unfriendly, see for example \cite{liao2016important} for more discussions.
Our learning model may serve as a building block for the pursuit of biologically plausible deep networks with backpropagation-style learning.

\begin{figure*}[t]
	\centering
	\small
	\begin{tabular}{cc}
		\includegraphics[scale=0.22]{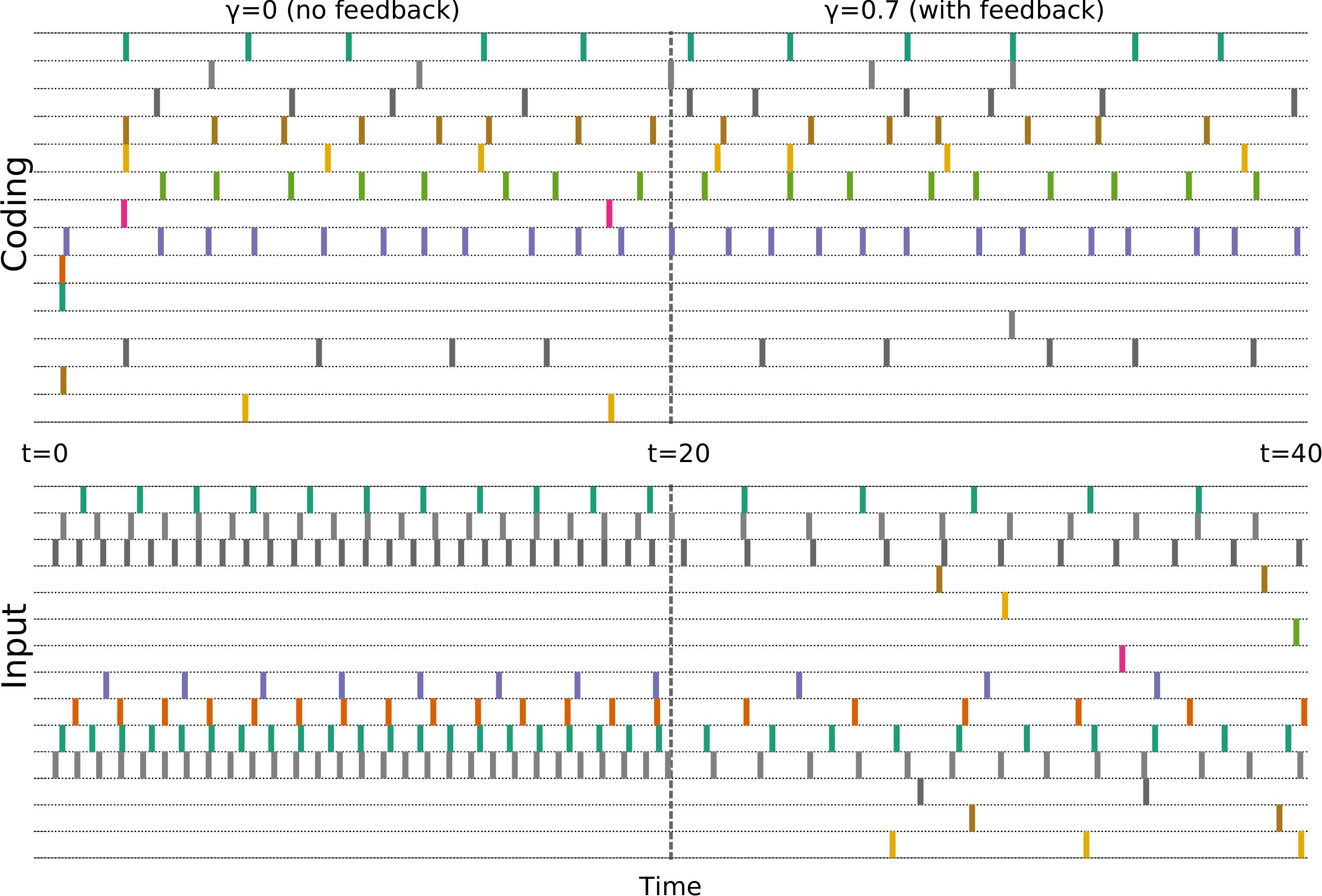} 
		&
		\includegraphics[scale=0.22]{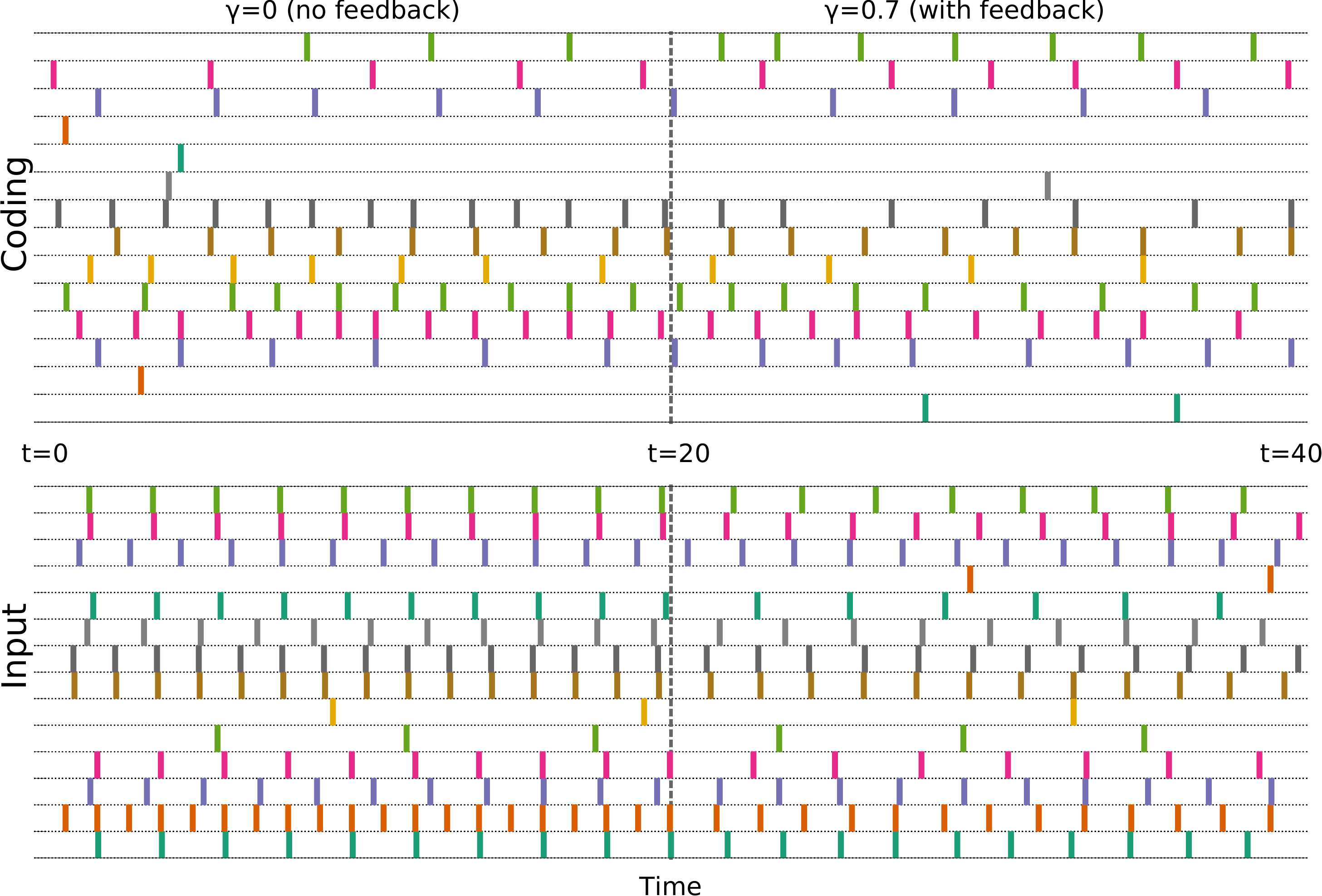} 
		\\
		(a) Random dictionary (training sample No.1)
		&
		(b) Learned dictionary (training sample No.99900)
	\end{tabular}
	\caption{Network spike patterns. In the figures, each row corresponds to one neuron, and the bars indicate the spike timings. 
		One notable difference between the left and right figures is in the spike patterns of the input neurons. 
		Before learning, significant perturbation in spike patterns can be observed starting at $t=20$ when the feedback is present.
		In contrast, little change in spike patterns is seen after learning. 
		Recall that the perturbation in spike rates reflects the reconstruction error.
		This shows the network is able to learn a proper dictionary that minimizes reconstruction error. Data is from learning with Dataset A; only a subset of the neurons are shown. }
	\label{fig:network_dynamics}
\end{figure*}


\section{Numerical Experiments}
\label{sec:experiments}
We examined the proposed learning algorithm using three datasets.
\textbf{Dataset A.} 100K randomly sampled $8\times 8$ patches from the grayscale Lena image to learn 256 atoms.  
\textbf{Dataset B.} 50K $28\times 28$ MNIST images \cite{lecun1998gradient} to learn 512 atoms.
\textbf{Dataset C.} 200K randomly sampled $16\times 16$ patches from whitened natural scenes \cite{olshausen1996emergence} to learn 1024 atoms. These are standard datasets in image processing (A), machine learning (B), and computational neuroscience (C).\footnote{For Dataset A and C, the patches are further subtracted by the means, normalized, and split into positive and negative channels to create non-negative inputs \cite{hoyer2004non}.}
For each input, the network is ran with $\gamma=0$ from $t=0$ to $t=20$ and with $\gamma=0.7$ from $t=20$ to $t=40$, both with a discrete time step of $1/32$.  
Note that although this time window of 20 is relatively small and yields a spike rate precision of only 0.05, we observed that it is sufficient for gradient calculation and dictionary learning purpose.

We explored two different connection weight initialization schemes. First, we initialize the weights to be fully consistent with respect to a random dictionary. 
Second, we initialized the weights to be asymmetric. In this case, we set $F^T$ and $B$ to be column-normalized random matrices and the entries of $H$ to be random values between [0, 1.5] with the diagonal set to 1.5.

\subsection{Network Dynamics}

We first show the spike patterns from a network with fully consistent initial weights in Figure \ref{fig:network_dynamics}.
It can be seen that the spike patterns quickly settle into a steady state, indicating that a small time window may suffice for spike rate calculations. 
Further, we can observe that feedback only perturbs the input neuron spike rates while keeping the coding neuron spike rates approximately the same, validating our results in Section \ref{sec:sparse_coding} and \ref{sec:dictionary_adjustment}.

\begin{figure}[t]
	\centering
	\setlength\tabcolsep{1.5pt}
\begin{tabular}{cc}
	\includegraphics[scale=0.33]{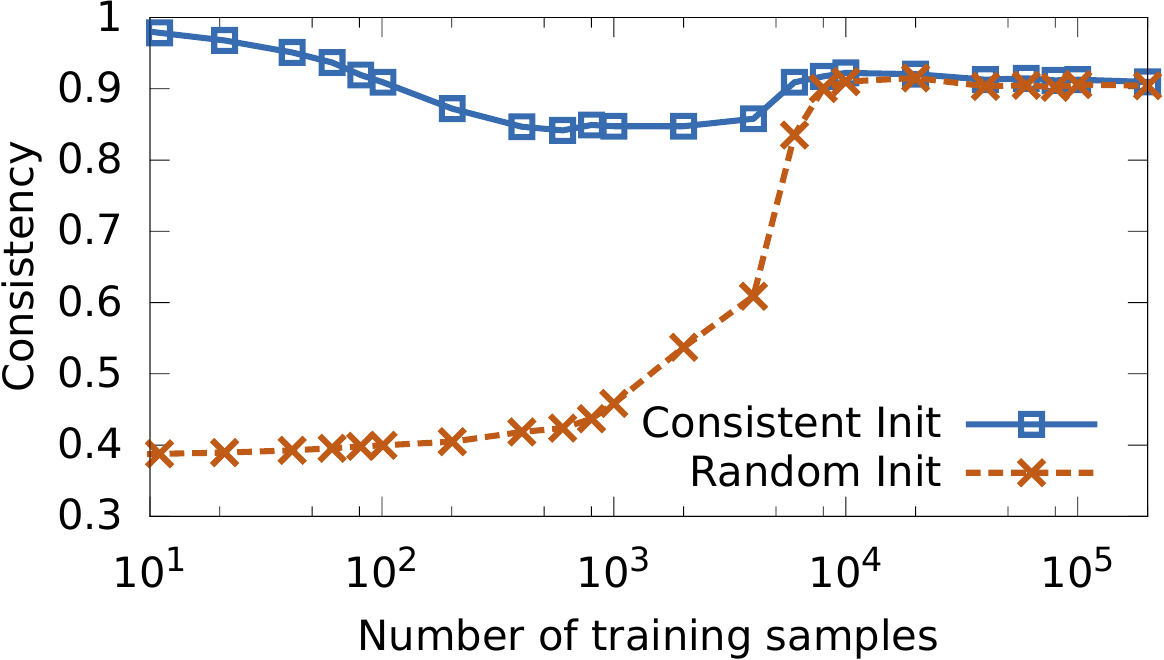}  
	&
	\includegraphics[scale=0.33]{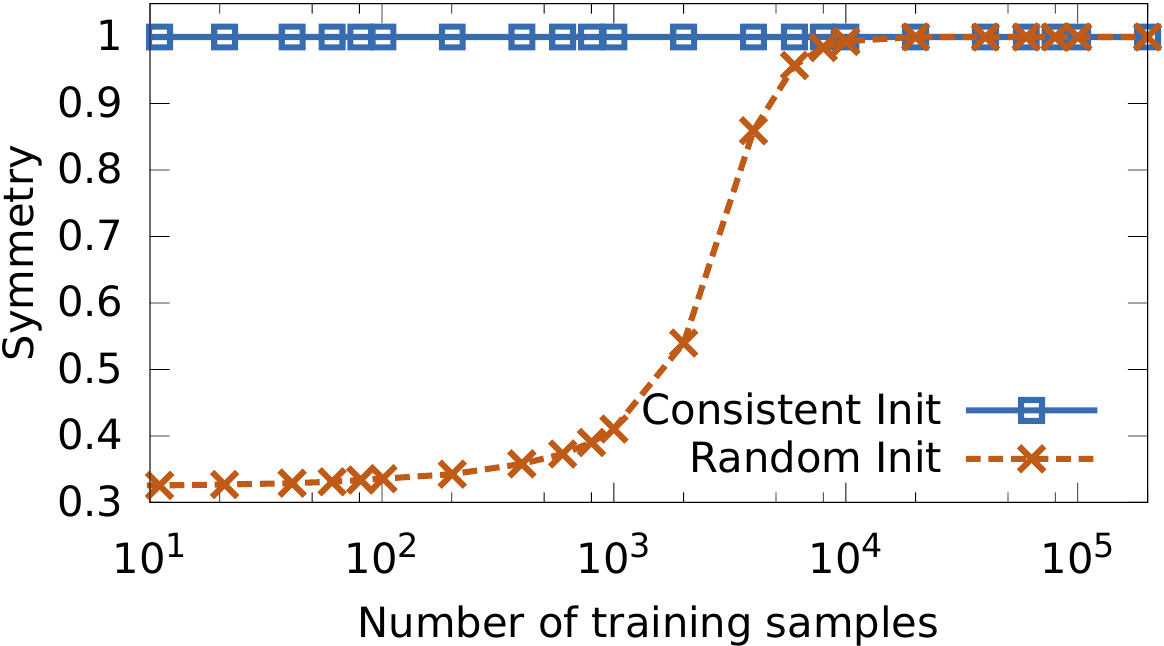}  
\end{tabular}	
	\caption{Network weight consistency and symmetry during learning. Consistency is measured as $1 - \left\| H-FB \right\|_F / \left\| H \right\|_F$. Symmetry is measured as the average normalized inner product between the $i$-th row of $F$ and the $i$-th column of $B$ for $i=1\hdots N$. Data is from learning with Dataset A.}
\label{fig:consistency}
\end{figure}

Another target the algorithm aims at is to approximately maintain the weight consistency $H \approx FB$ during learning. 
Figure \ref{fig:consistency} shows that this is indeed the case. Note that our learning rule acts as a catch-up correction, and so an exact consistency cannot be achieved.
An interesting observation is that as learning proceeds, weight consistency becomes easier to maintain as the dictionary gradually converges.

Although we have limited theoretical understanding for networks with random initial weights, Figure \ref{fig:consistency} shows that our learning procedure can automatically discover consistent and symmetric weights with respect to a single global dictionary. 
This is especially interesting given that the neurons only learn with local information. No neuron has a global picture of the network weights.



\begin{figure*}[t]
	\centering
	\begin{tabular}{ccc}
		\includegraphics[scale=0.37]{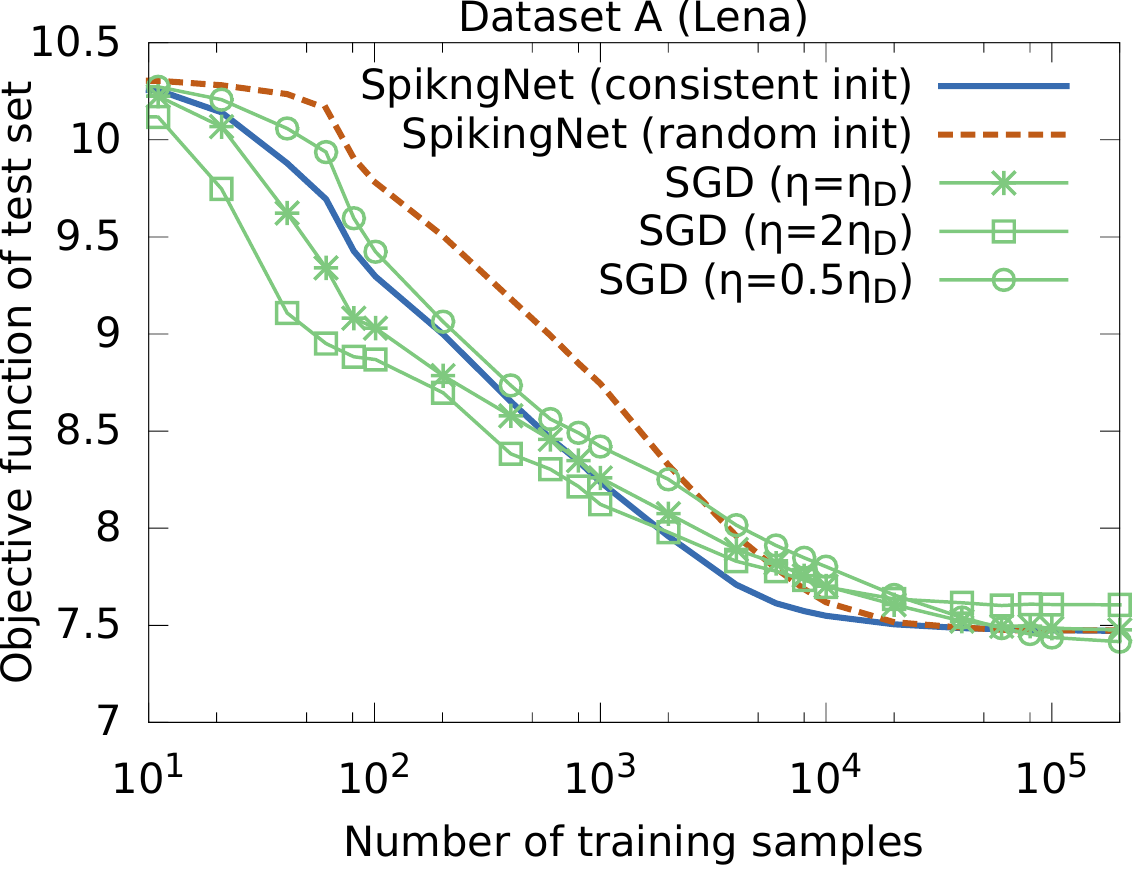} &
		\includegraphics[scale=0.36]{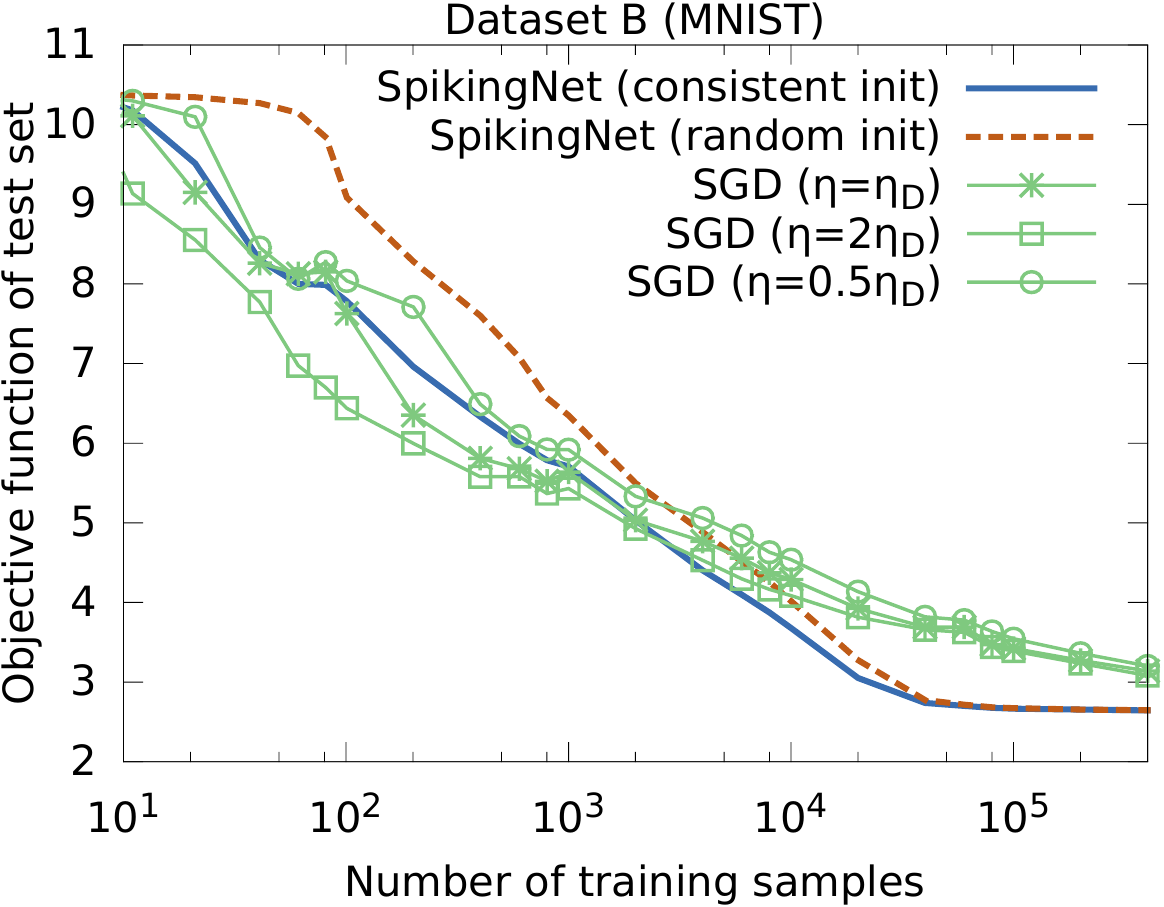} & 
		\includegraphics[scale=0.36]{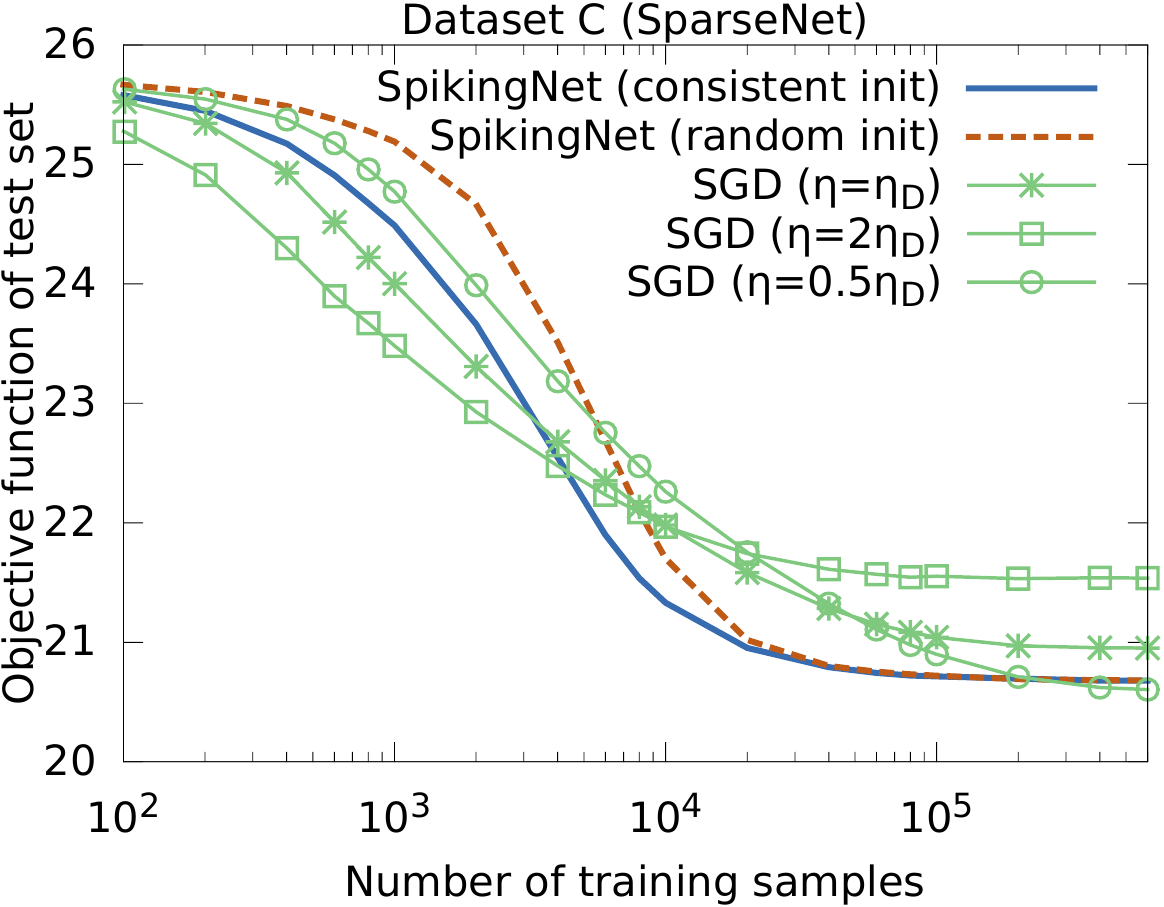} 
	\end{tabular}		
	\caption{Comparison of convergence of learning with dynamical neural network and SGD. 
	}
	\label{fig:convergence}
\end{figure*}

\subsection{Convergence of Dictionary Learning}

The learning problem is non-convex, 
and hence it is important that our proposed algorithm can find a satisfying local minimum. 
We compare the convergence of spiking networks with the standard stochastic gradient descent (SGD) method with 
the unit atom norm constraint.
For simplicity, both algorithms use a batch size of 1 for gradient calculations. 
The quality of the learned dictionary $D=F^T$ is measured using a separate test set of 10K samples to calculate a surrogate dictionary learning objective \cite{mairal2009online}. 
For a fair comparison, the weight decay parameters in spiking networks are chosen so that the average atom norms converge to approximately one.

Figure \ref{fig:convergence} shows that our algorithm indeed converges and can obtain a solution of similar, if not better, objective function values to SGD consistently across the datasets. 
Surprisingly, our algorithm can even reach a better solution with fewer training samples, while SGD can be stuck at a poor local minimum especially when the dictionary is large.
This can be attributed to the $\ell_1$-norm reweighting heuristic that encourages 
more dictionary atoms to be actively updated during learning.
Finally, we observe that a network initialized with random non-symmetric weights still
manages to reach objective function values comparable to those initialized with symmetric weights,
albeit with slower convergence due to less accurate gradients.
From Figure \ref{fig:consistency}, we see the network weights are not symmetric before $10^4$ samples for Dataset A.
On the other hand, from Figure \ref{fig:convergence} the network can already improve the dictionary before $10^4$ samples, showing that perfectly symmetric weights are not necessary for learning to proceed.

\section{Conclusion}
We have presented a dynamical neural network formulation that can learn dictionaries for sparse representations. 
Our work represents a significant step forward that it not only provides a link between the well-established dictionary learning problem and dynamical neural networks, but also demonstrates 
the contrastive learning approach to be a fruitful direction.
We believe there is still much to be explored in dynamical neural networks.
In particular, learning in such networks respects data locality and therefore 
has the unique potential, especially with spiking neurons, to enable low-power, high-throughput training with massively parallel architectures.


\newpage
\appendix
\appendixpage

\section{Detailed Description of Proposed Network Structure}
We propose a novel network topology with feedback shown in Figure~\ref{fig:networks}(b). 
The figure shows two ``layers'' of neurons.
The lower layer consists of $M$ neurons we call input neurons, $n_{I,i}$
for $i=1,2,\ldots,M$; the upper layer consists of $N$ neurons we call
coding neurons $n_{C,i}$ for $i=1,2,\ldots,N$. 

Each coding neuron $n_{C,i}$ receives excitatory signals from all the
input neurons $n_{I,j}$ with a weight of $F_{ij} \ge 0$. That is, each
coding neuron has a row of the matrix $F \in \reals_{\ge 0}^{N\times M}$.
In addition, neuron $n_{C,i}$ receives inhibitory signals from all other
coding neurons $n_{C,j}$ with weight $-W_{ij} \le 0$. 
$W$ denotes
this matrix of weights: $W\in\reals_{\ge 0}^{N\times N}$ and
${\rm diag}(W) = \vvzero$.
The firing thresholds are 
$\thresvec = [\thres_1,\thres_2,\ldots,\thres_N]^T$ and the matrix
$W+\Thres$, $\Thres = {\rm diag}(\thresvec)$, appears often and will
denote it as $H \bydef W + \Thres$.
Each neuron $n_{C,i}$ also receives a constant negative bias of 
$-(1-\gamma)\lambda_1 s_i$ where $0 \le \gamma < 1$ is an important parameter
that will be varied during the learning process to be detailed momentarily.

Each input neuron $n_{I,i}$, $i=1,2,\ldots,M$, with firing threshold fixed
to be 1, receives a bias of $(1-\gamma)x_i$. Typically
$x_i$ corresponds to the $i$-th pixel value of an input image in question
during the learning process. In addition, it receives excitatory spikes
from each of the coding neurons with weights $\gamma B_{ij} \ge 0$.
That is each input neuron has a row of the matrix $B\in\reals_{\ge 0}^{M\times N}$.
These excitatory signals from the coding neurons constitute 
the crucial feedback mechanism we devised here that enables dictionary learning.

\section{Proof of Theorems}

\subsection{Theorem 1: SNN dynamics, trajectory, and limit points}

In the simplest case when none of the neurons are inter-connected and $\poten_i(0) < \thres_i$
for all $i$, then $\curr_i(t) = \bias_i$ for all $i$ and all $t \ge 0$. Hence those
neurons $n_i$ with $\bias_i > 0$ produces a spike train of constant inter-spike interval 
of $\thres_i/\bias_i$; those neurons with $\bias_i \le 0$ will have no spiking activities.
When however the neurons are inter-connected, the dynamics becomes non-trivial. It turns
out that one can so describe the dynamics mathematically that useful properties
related to the current and spike train can be derived. Consequently, a network of spiking
neurons can be configured to help solve certain practical problems.

Given a system of $N$ neurons $n_i$, $i=1,2,\ldots,N$, we use vector notations
$\currvec(t)$ and $\spiketrainvec(t)$ to denote the $N$ currents and spike trains.
The vector $\biasvec$ and $\thresvec$ are the input biases and firing thresholds.
The convolution $(\alpha\ast\spiketrainvec)(t)$ is the $N$-vector whose
$i$-th component is $(\alpha\ast\spiketrain_i)(t)$. For simplicity, we consider
only $\tau=1$ throughout the paper. Thus $\alpha(t) = e^{-t}$ for $t\ge 0$ and 0 otherwise. 
Equation~\ref{eqn:soma_current}
in vector form is
\begin{equation}
\label{eqn:vector_soma_current}
\currvec(t) = \biasvec + W (\alpha\ast\spiketrainvec)(t)
\end{equation}
where $W\in\reals^{N\times N}$ and $W_{ii}=0$, encodes the inhibitory/excitatory 
connections among the neurons. 

Because 
$\frac{d}{dt} (\alpha\ast\spiketrain)(t) = \spiketrain(t)-(\alpha\ast\spiketrain)(t)$,
we have
\begin{equation}
\label{eqn:appx_current_dot}
\dot{\currvec}(t) = \biasvec - \currvec(t) + W\cdot\spiketrainvec(t).
\end{equation}
Filtering Equation~\ref{eqn:appx_current_dot} yields
\begin{align}
\dot{\avgcurrvec}(t) &= 
\biasvec - \avgcurrvec(t) + W\,\spikeratevec(t) + (\currvec(0)-\avgcurrvec(t))/t \nonumber \\
\avgcurrvec(t)-\Thres \,\spikeratevec(t) &=
\biasvec + (W-\Thres)\,\spikeratevec(t) \nonumber \\
 &\phantom{kk}
   + (\currvec(0)-\avgcurrvec(t))/t - \dot{\avgcurrvec}(t)
\label{eqn:avgcurr_spikerate_dynamics}
\end{align}
where $\Thres = {\rm diag}(\thresvec)$. Theorem~\ref{thm:basic_limits} has been established
previously in~\cite{tang2016sparse} in a slightly different form. 
We attach the proof consistent to our notations below for completeness.
It is established under the following assumptions: 
\begin{itemize}
	\item The currents of all neurons remain bounded from above,
	$\Maxnorm{\currvec(t)}\le B$ for all $t\ge 0$ for some $B > 0$. 
	This implies no neuron can spike arbitrarily fast, and the fact that neurons cannot spike arbitrarily rapidly implies the currents are bounded from below
	as well
	\item There is a positive number $r > 0$ such that whenever the numbers
	$t_{i,k}$ and $t_{i,k+1}$ exist, $t_{i,k+1} - t_{i,k} \le 1/r$. This assumption says that unless a neuron stop spiking althogether after a certain time, the duration between consecutive spike cannot become arbitrarily long.
\end{itemize}

\begin{appxthm}
\label{thm:basic_limits}
As $t\tendsto \infty$,
$\dot{\avgcurrvec}(t)$, 
$\frac{1}{t}(\currvec(0)-\avgcurrvec(t))$ and
$\max(\avgcurrvec(t),\vvzero)-\Thres\,\spikeratevec(t)$
all converge to $\vvzero$. 
\end{appxthm}
\begin{myproof}
	Let 
	$$
	\ccA = \{\, i \suchthat \mbox{neuron-$i$ spikes infinitely often}\, \}
	$$
	($\ccA$ stands for ``active''), and
	$$
	\ccI = \{\, i \suchthat \mbox{neuron-$i$ stop spiking after a finite time} \,\}
	$$
	($\ccI$ stands for ``inactive'').
	First consider $i \in \ccI$. Let $\spt{i}{k}$ be the time of the final spike. For any
	$t > \spt{i}{k}$,
	\begin{eqnarray*}
		\avgcurr_i(t) 
		& = & \frac{1}{t}\int_0^{\spt{i}{k}} \curr_i(s)\,ds + 
		\frac{1}{t}\int_{\spt{i}{k}}^t \curr_i(s)\,ds \\
		& = & \frac{1}{t}\int_0^{\spt{i}{k}} \curr_i(s)\,ds + 
		\frac{1}{t}\poten_i(t) \\
		& = & \thres_i\spikerate_i(t) + \frac{1}{t}\poten_i(t)
	\end{eqnarray*}
	Note that $\poten_i(t) \le \thres_i$ always. If $\poten_i(t) \ge 0$, then
	$$
	0 \le \max(\avgcurr_i(t),\, 0)  - \thres_i \spikerate_i(t) \le \thres_i/t.
	$$
	If $\poten_i(t) < 0$, 
	$$
	-\thres_i\spikerate_i(t) \le \max(\avgcurr_i(t),\, 0) - \thres_i\spikerate_i(t) \le 0.
	$$
	Since $i\in\ccI$, $\spikerate_i(t) \tendsto 0$ obviously. Thus
	$$
	\max(\avgcurr_i(t),\,0) - \thres_i\spikerate_i(t)  \tendsto 0.
	$$
	
	\VS
	Consider the case of $i\in\ccA$. 
	For any $t > 0$, let $\spt{i}{k}$ be the largest spike time
	that is no bigger than $t$. Because $i\in\ccA$, $\spt{i}{k} \tendsto \infty$ as $t\tendsto\infty$.
	\begin{eqnarray*}
		\avgcurr_i(t) 
		& = & \frac{1}{t}\int_0^{\spt{i}{k}} \curr_i(s)\,ds + 
		\frac{1}{t}\int_{\spt{i}{k}}^t \curr_i(s)\,ds  \\
		& = & \thres_i\spikerate_i(t) + 
		\frac{1}{t}\int_{\spt{i}{k}}^t \curr_i(s)\,ds.
	\end{eqnarray*}
	Furthermore, note that because of the assumption $\spt{i}{k+1}-\spt{i}{k} \le 1/r$ always,
	where $r >0$, $\lim\inf \spikerate_i(t) \ge r$. In otherwords, there is a time $T$ large enough
	such that $\spikerate_i(t) \ge r/2$ for all $i\in\ccA$ and $t\ge T$.
	Moreover, $0 \le t - \spt{i}{k} \le \spt{i}{k+1} - \spt{i}{k} \le 1/r$ and 
	$\curr_i(t) \in [B_-, B_+]$. Thus 
	$$
	\frac{1}{t}\int_{\spt{i}{k}}^t \curr_i(s)\,ds 
	\in \frac{1}{t} \, [B_-, B_+] / r \tendsto 0.
	$$
	When this term is eventually smaller in magnitude than $\thres_i\spikerate_i(t)$, we have
	$$
	\avgcurr_i(t) - \thres_i\spikerate_i(t)  \tendsto 0.
	$$
	or equivalently,
	$$
	\max(\avgcurr_i(t),\,0) - \thres_i\spikerate_i(t)  \tendsto 0.
	$$
	\qed
\end{myproof}

Applying Theorem~\ref{thm:basic_limits} to Equation~\ref{eqn:avgcurr_spikerate_dynamics}
yields the following.
\begin{appxthm}
\label{thm:trajectory_large_time}
Given any $\epsilon > 0$, there exists $T > 0$ such that for all $t>T$,
\[
\Maxnorm{ (\avgcurrvec(t)-\Thres\,\spikeratevec(t)) -
(\biasvec + (W-\Thres)\,\spikeratevec(t)) } < \epsilon.
\]
\end{appxthm}
The following theorem characterizes limit points of the trajectory $(\avgcurrvec(t),\spikeratevec(t))$.
Recall that $(\avgcurrvec^*,\spikeratevec^*)$ is a limit point if given any $\epsilon > 0$,
there exists a time $T > 0$ large enough such that $(\avgcurrvec(T),\spikeratevec(T))$ is within
$\epsilon$ to $(\avgcurrvec^*,\spikeratevec^*)$.
\begin{appxthm}
	\label{thm:limit_points}
	Given any limit point $(\avgcurrvec^*,\spikeratevec^*)$, 
	we must have 
	$\biasvec + (W - \Thres)\,\spikeratevec^* \le \vvzero$, $\spikeratevec^*\,\odot\, (\biasvec + (W - \Thres)\,\spikeratevec^*) = \vvzero$ and 
	$\spikeratevec^* \ge 0$,
	where $\odot$ is the elementwise product.
\end{appxthm}
\begin{myproof}
Theorem~\ref{thm:basic_limits} shows that
$\avgcurrvec^* - \Thres\,\spikeratevec^* \le \vvzero$ and the
elementwise product
$\spikeratevec^*\,\odot\, (\avgcurrvec^* - \Thres\,\spikeratevec^*) = \vvzero$.
But Theorem~\ref{thm:trajectory_large_time} shows that
$
\avgcurrvec^* - \Thres\,\spikeratevec^*  = 
\biasvec + (W - \Thres)\,\spikeratevec^*
$
and the theorem here is established. \qed
\end{myproof}

\subsection{Non-negative sparse coding by spiking neural networks}
\label{sec:non-negative-sparse-coding}

\begin{figure}[t]
	\center
	\setlength\tabcolsep{1.5pt}
	\includegraphics[scale=0.35]{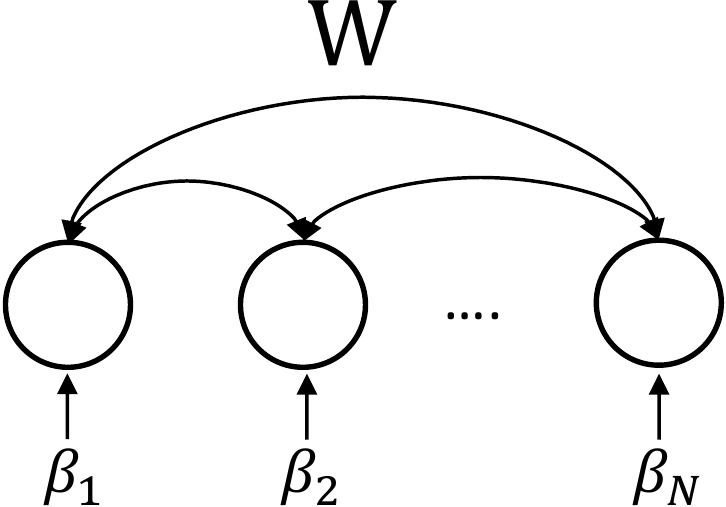}  
	\caption{A 1-layer LCA network for sparse coding.
	}
	\label{fig:appx_networks}
\end{figure}

Given a non-negative dictionary $D\in\reals_{\ge 0}^{M\times N}$,
a positive 
scaling vector $\vvs = [s_1,s_2,\ldots,s_N]^T\in\reals_{>0}^N$ and
an image $\vvx\in\reals_{\ge 0}^M$,
the non-negative sparse coding problem can be formulated as
\begin{equation}
\vva^* = \argmin_\mathbf{\vva \geq \vvzero} 
l\left(\vva\right), 
l(\vva)=
\frac{1}{2}\Enormsq{ \vvx - D\,\vva } +
\lambda\,\Onenorm{ S\,\vva }
\label{eqn:nn_lasso}
\end{equation}
where $S={\rm diag}(\vvs)$.
Using the well-known KKT condition in optimization theory,
see for example \cite{BoydVandenberghe04},
$\vva^*$ is an optimal solution iff there exists $\vve^*\in\reals^N$ such that
all of the following hold:
\begin{align}
\left\{\begin{matrix}
&\vvzero \in \partial l(\vva^*) - \vve^* & (\textrm{stationarity}) \\
&\vve^* \odot \vva^* = \vvzero & (\textrm{complementarity})\\
&\vva^* \ge \vvzero, \vve^* \ge \vvzero & (\textrm{feasibility})
\end{matrix}\right.
\label{eq:kkt_sc}
\end{align}
where $\partial l$ is the generalized gradient of $l$. 
Note that the generalized gradient $\partial l(\vva)$ is
$D^T D\vva - D^T \vvx + \lambda \vvs \odot \partial \Onenorm{\vva}$
and that $\partial |a_i| = 1$ when $a_i > 0$ and 
equals the interval $[-1,1]$ when $a_i = 0$.
Straightforward derivation then shows that
$\vva^*$ is an optimal solution iff
\begin{equation}
\label{eqn:applied_kkt}
\begin{aligned}
D^T\,\vvx - \lambda\,\vvs - D^T D \vva^* \ & \le  \vvzero\ ;{\rm and}\\
\vva^*\,\odot\, 
(D^T\,\vvx - \lambda\,\vvs - D^T D \vva^*) & = \vvzero.
\end{aligned}
\end{equation}

We now configure a $N$-neuron system depicted in Figure~\ref{fig:appx_networks}
so as to solve Equation~\ref{eqn:nn_lasso}.
Set $\thresvec = {\rm diag}(D^T D)$ as the firing thresholds and set
$\biasvec = D^T\vvx - \lambda\vvs$ as the bias.
Define the inhibition matrix to be $-(D^T D - \Thres)$,
$\Thres = {\rm diag}(\thresvec)$. Thus neuron-$j$ inhibits
neuron-$i$ with weight $-\vvd_i^T \vvd_j \le 0$. In this configuration,
it is easy to establish $\Maxnorm{\currvec(t)} \le C$ for all 
$t\ge 0$ for some $C > 0$ as all connections are inhibitions. From 
Theorem~\ref{thm:limit_points}, any limit point $(\avgcurrvec^*,\spikeratevec^*)$
of the trajectory $(\avgcurrvec(t),\spikeratevec(t))$ satisfies 
$\biasvec+(W-\Thres)\spikeratevec^* \le \vvzero$ and
$\spikeratevec^*\,\odot\, (\biasvec+(W-\Thres)\spikeratevec^* )  = \vvzero$.
But $\biasvec = D^T\vvx - \lambda\,\vvs$ and
$W-\Thres = -D^T D$. Thus $\spikeratevec^*$ solves 
Equation~\ref{eqn:nn_lasso}. And in particular, if the solution to
Equation~\ref{eqn:nn_lasso} is unique, the trajectory can only have one
limit point, which means in fact the trajectory converges to the sparse
coding solution.
This result can be easily extended to the network in Figure~\ref{fig:networks}(a) by expanding the bias into another layer of input neuron with $F=D^T$.

\subsection{Theorem 2: sparse coding with feedback perturbation}

Equation~\ref{eqn:limits_of_slacks}, reflecting the structure of the coding and input neurons,
takes the form:
\begin{equation}
\label{eqn:block_eqn}
\begin{aligned}
\left[
\begin{matrix}
\vv{e}_\gamma(t) \\
\vv{f}_\gamma(t)
\end{matrix}
\right]
& \bydef
\left[
\begin{matrix}
\vv{u}_\gamma(t)-\Thres\vv{a}_\gamma(t) \\
\vv{v}_\gamma(t)-\vv{b}_\gamma(t)
\end{matrix}
\right]  =
\left[
\begin{matrix}
-(1-\gamma)\lambda_1\vv{s} \\
(1-\gamma)\vv{x}
\end{matrix}
\right]
+
\left[
\begin{matrix}
-H &  F \\
\gamma B   & -I 
\end{matrix}
\right]
\,
\left[
\begin{matrix}
\vv{a}_\gamma(t)\\
\vv{b}_\gamma(t)
\end{matrix}
\right] 
+
\bb{\Delta}(t)
\end{aligned}
\end{equation}
$(\vv{u}(t),\vv{v}(t))$ and
$(\vv{a}(t),\vv{b}(t))$ denote the average currents and
spike rates for the coding and input neurons, respectively, and
$H \bydef W + \Thres$. 
Note that
$\max(\vv{u}_\gamma(t),\vvzero)-\Thres\vv{a}_\gamma(t)$,
$\max(\vv{v}_\gamma(t),\vvzero)-\vv{b}_\gamma(t)$ and
$\bb{\Delta}(t)$ all converge to $\vvzero$ as
$t \rightarrow \infty$.

\begin{appxthm}
	\label{thm:perturbation}
	Consider the configuration $F^T=B=D$ and $\gamma\in[0,1)$. Suppose the
	soma currents and thus spike rates $\Maxnorm{\vv{a}_\gamma(t)}$ are bounded.
	Let $H = D^TD + (\lambda_1\oneminus{\gamma})^{-1}\Delta_H$, $\oneminus{\gamma} = 1-\gamma$, be such that
	$4\Onenorm{\Delta_H}\Maxnorm{\vv{a}_\gamma(t)} < \min\{s_i\}$.
	Then, for any $\epsilon > 0$ there is $T > 0$ such that for all $t>T$,
	$\Maxnorm{\vv{a}_\gamma(t)-\perturbed{\vv{a}}(t)} < \epsilon$ and 
	$\perturbed{\vv{a}}(t)$ solves Equation~\ref{eqn:nn_lasso} with
	$S$ replaced by $\perturbed{S}$ where
	$\Maxnorm{S-\perturbed{S}} < \min\{s_i\}/2$.
\end{appxthm}
\begin{myproof}
Consider $\tau > 0$ and define 
the vectors $\perturbed{\vv{a}}(t)$ and
$\perturbed{\vv{u}}(t)$ for $t\ge 0$ by each of their components:
\[
(\perturbed{a}_i(t),\perturbed{u}_i(t)) = \left\{
\begin{array}{l l}
(a_{\gamma,i}(t), \thres_i a_{\gamma,i}(t))   & \hbox{if $a_{\gamma,i}(t) \ge \tau$,} \\
(0,\phantom{kk}  \min(u_{\gamma,i}(t),0))     & \hbox{otherwise,}
\end{array}
\right.
\]
where $\thresvec$ is the diagonal of $D^T D$. Denote the perturbations
$\bb{\Delta}_a(t) \bydef \perturbed{\vv{a}}(t)-\vv{a}_\gamma(t)$,
$\bb{\Delta}_u(t) \bydef \perturbed{\vv{u}}(t)-\vv{u}_\gamma(t)$,
$\perturbed{\vv{e}}_\gamma(t) \bydef
\perturbed{\vv{u}}(t) - \Thres\perturbed{\vv{a}}(t)$, and
$\bb{\Delta}_e(t) \bydef \perturbed{\vv{e}}(t)-\vv{e}_\gamma(t)$.
This construction of $\perturbed{\vv{u}}(t)$ and $\perturbed{\vv{a}}(t)$
ensures $\Maxnorm{\bb{\Delta}_a(t)} < \epsilon$,
$\perturbed{\vv{e}}(t)\le\vvzero$, and
$\perturbed{\vv{e}}(t)\odot\perturbed{\vv{a}}(t) = \vvzero$.
Recall that $\max(\vv{u}_\gamma(t),\vvzero)-\Thres\vv{a}_\gamma(t)\rightarrow\vvzero$
(Theorem~\ref{thm:basic_limits}); 
thus $\Maxnorm{\bb{\Delta}_u(t)} < 2\tau$ at $t$ large enough.

Next, observe that $\vv{v}_\gamma(t)\ge \vvzero$ always $\bb{\nu}_\gamma(t)\ge\vvzero$ always,
for any setting $\gamma$ in $[0,1)$. Thus Theorem~\ref{thm:basic_limits} implies
\begin{equation}
\label{eqn:limit_of_b}
\vv{b}_\gamma(t) - [ (1-\gamma)\vv{x}+\gamma B \vv{a}_\gamma(t) ] \rightarrow \vvzero
\end{equation} as
$t\rightarrow \infty$.
From Equation~\ref{eqn:block_eqn}, this implies that
\[
\bb{e}_\gamma(t) = 
\oneminus{\gamma}(D^T\vv{x}-\lambda_1\vv{s}-D^T D \vv{a}_\gamma(t) - \lambda_1 \Delta_H\vv{a}_\gamma(t) + 
\bb{\Delta}(t))
\]
for some $\bb{\Delta}(t)$ where $\Maxnorm{\bb{\Delta}(t)}\rightarrow 0$. Thus
\[
(\oneminus{\gamma})^{-1} \perturbed{\vv{e}}(t) =
D^T \vv{x} - \lambda_1 \perturbed{\vv{s}} - D^T D \perturbed{\vv{a}}(t)
\]
where $\perturbed{\vv{s}} = \vv{s}-(\bb{\eta}(t)+\bb{\zeta}(t))$,
$\bb{\eta}(t)=\Delta_H \vv{a}_\gamma(t)$ and
$\bb{\zeta}(t) = 
\lambda_1^{-1}(D^TD \bb{\Delta}_a(t) + \bb{\Delta}_e(t)/\oneminus{\gamma} + \bb{\Delta}(t))$.
By assumption on $\Delta_H$, 
$\vv{s}-\bb{\eta}(t) > (3/4)\vv{s} > \vvzero$.
Moreover,$\Maxnorm{\bb{\zeta}(t)}$ can be made arbitrarily small by 
taking $t$ and $1/\tau$ large enough. Thus there exist 
$\tau, T > 0$ such that for all $t > T$, $\Maxnorm{\bb{\Delta}_a(t)} < \epsilon$ and
$\Maxnorm{\perturbed{\vv{s}}(t) - \vv{s}} < \min\{s_i\}/2$, implying in particular 
$\perturbed{\vv{s}}(t) > \vv{s}/2 > \vvzero$. Finally, note that
\[
\perturbed{\vv{a}}(t)\ge\vvzero, \phantom{kk}
(\oneminus{\gamma})^{-1}\perturbed{\vv{e}}(t)\le\vvzero, \phantom{kk}
(\oneminus{\gamma})^{-1}\perturbed{\vv{e}}(t)\odot
\perturbed{\vv{a}}(t) = \vvzero,
\]
which shows (recall Equation~\ref{eqn:applied_kkt}) that
$\perturbed{\vv{a}}(t)$ solves Equation~\ref{eqn:nn_lasso} with
$S$ replaced by $\perturbed{S}$ and the proof is complete. \qed
\end{myproof}

At present, we cannot establish a priori that the currents stay bounded
when $\gamma > 0$. Nevertheless, the theorem is applicable in practice
as long as the observed currents stay bounded by some $C$ for $0\le t \le T$
and $C/T$ is small enough. See Section~\ref{sec:discussions} for further
comments.

\subsection{Theorem 3: gradient calculations from contrastive learning}

\begin{appxthm}
	\label{thm:approx_for_DA}
	Given any $\epsilon > 0$, there is a $T > 0$ such that for all $t, t' > T$,
	\begin{align}
	&
	\Maxnorm{ 
		\kappa(B\vv{a}_\kappa(t') - \vv{x}) - 
		(\vv{b}_\kappa(t')-\vv{b}_0(t))  
	} < \epsilon,
	\label{eqn:appx_limiting_states_b} \\
	&
	\| 
	\oneminus{\kappa} H(\vv{a}_0(t)-\vv{a}_\kappa(t')) -  
	\kappa(H - FB)\vv{a}_\kappa(t')  + 
	(\oneminus{\kappa} \vv{e}_0(t) - \vv{e}_\kappa(t') ) \|_\infty < \epsilon.
	\label{eqn:appx_limiting_states_e}
	\end{align}
\end{appxthm}
\begin{myproof}
	Equation~\ref{eqn:limit_of_b} implies that
	\[
	\kappa(B\vv{a}_\kappa(t') - \vv{x}) - 
	(\vv{b}_\kappa(t')-\vv{b}_0(t))  
	\rightarrow \vvzero
	\quad \hbox{as $t,t'\rightarrow\infty$},
	\]
	establishing Equation~\ref{eqn:appx_limiting_states_b}.
	From Equations~\ref{eqn:block_eqn} and~\ref{eqn:limit_of_b}
	\begin{align*}
	&
	-\oneminus{\kappa}\lambda\vv{s} - \oneminus{\kappa} H\vv{a}_0(t) + \oneminus{\kappa}F\vv{x}
	- \oneminus{\kappa}\vv{e}_0(t) \rightarrow \vvzero, \;{\rm and}, \\
	& 
	-\oneminus{\kappa}\lambda\vv{s} - H\vv{a}_\kappa(t) + \oneminus{\kappa}F\vv{x} + \kappa FB \vv{a}_\kappa(t) 
	- \vv{e}_\kappa(t) \rightarrow \vvzero.
	\end{align*}
	Equation~\ref{eqn:appx_limiting_states_e} thus follows. \qed
\end{myproof}
\section{Comparisons with Prior Work}

\subsection{Comparisons of dynamical neural networks}

Table \ref{tab:dynn_compare} provides a comparison between the development of three types of dynamical neural networks: Hopfield network, Boltzmann machine, and sparse coding network.
Text in boldface indicates the new results established in this work.

\begin{table}[h]
\begin{tabular}{| p{1.5cm} | p{4.7cm} | p{4.5cm} | p{4.5cm} |}
	\hline
	& Hopfield network & Boltzmann machine & Sparse coding network
	\\ \hline
	Neuron model &
	Binary \cite{hopfield82} or continuous \cite{hopfield84} & 
	Binary \cite{ackley1985learning} or continuous (for visible units) \cite{freund1992unsupervised} & 
	Continuous \cite{rozell2008sparse} or spiking \cite{shapero2014optimal} 
	\\ \hline
	Activation &
	Binary: Thresholding \hphantom{kkk} 
	Cont.: Any bounded, differentiable, strictly increasing function&
	Logistic&
	Rectified linear

	\\ \hline	
	Topology &
	Arbitrary symmetric bidirectional connections &
	BM: Arbitrary symmetric bidirectional connections \hphantom{kkk} 
	RBM \cite{hinton2006fast}: Two-layer with symmetric forward/backward &
	Two-layer with feedforward, lateral, and \textbf{feedback} connections

	\\ \hline
	Learning &
	Binary: Hebbian rule\hphantom{kkk} 
	Cont.: contrastive learning\cite{movellan1991contrastive}
	& BM: contrastive learning \hphantom{kkk} 
	RBM: contrastive divergence
	& \textbf{Contrastive learning with weight consistency}
	\\ \hline
    Limit point &
    Many local minimum & Many local minimum & Likely unique \cite{bruckstein2008uniqueness}
    \\ \hline
	Usage & Associative memory, constraint satisfaction problem & Generative model, constraint satisfaction problem & Representation learning with sparse prior, image denoising and super-resolution, compressive sensing
	\\ \hline
\end{tabular}
\caption{Comparison between dynamical neural networks.}
\label{tab:dynn_compare}
\end{table}

\subsection{Comparisons of dictionary learning networks}

As we discussed in Section~\ref{sec:related_work}, there are several prior work that qualitatively demonstrate dictionary learning results in dynamical neural networks. 
The prior work~\cite{foldiak1990forming,zylberberg2011sparse,brito2016nonlinear,hu2014hebbian,seung2017correlation,vertechi2014unsupervised,brendel2017learning} employ a feedforward-only network topology as shown in Figure~\ref{fig:networks}(a), and are unable to compute the true gradient for dictionary learning from local information. These work hence rely on additional heuristic or assumptions on input data for learning to work. In contrast, we propose to introduce feedback connections as shown in Figure~\ref{fig:networks}(b), which allows us to solve the fundamental problem of estimating the true gradient.
Recall the dictionary learning objective function (Equation~\ref{eqn:dictionary_learning} in the main text)
\begin{equation}
\label{eqn:appx_dictionary_learning}
\argmin_{\ssup{\vva}{p} \ge \vvzero,
	D \ge \vvzero}
\sum_{p=1}^P l(D,\ssup{\vvx}{p},\ssup{\vva}{p}), \textrm{\phantom{k}}
l(D,\vvx,\vva) = \frac{1}{2}\Enormsq{\vvx - D\vva} + \lambda_1 \Onenorm{S\vva}
+ \frac{\lambda_2}{2} \Fnormsq{D},
\end{equation}
and the true stochastic gradient of the learning problem is (Equation~\ref{eqn:update_D} in the main text)
\begin{equation}
\label{eqn:appx_update_D}
D^{({\rm new})} \leftarrow
D - \eta \left((D\vv{a}-\vv{x})\vv{a}^T + \lambda_2 D \right).
\end{equation}
Here we provide a detailed discussion on the difference and limitations of prior work.

The first line of work is the so-called Hebbian/anti-Hebbian network~\cite{foldiak1990forming,zylberberg2011sparse,brito2016nonlinear,hu2014hebbian,seung2017correlation}.
The principle of learning in these work is to apply Hebbian learning to learn excitatory feedforward weights (strengthen the excitatory weights if both input and coding neurons have strong activations) and anti-Hebbian learning for inhibitory lateral weights (strengthen the inhibitory weights if both coding neurons have strong activations).
Due to heuristic nature of the learning rules, it is unclear whether this approach can solve the dictionary learning problem in Equation~\ref{eqn:appx_dictionary_learning}.
\cite{zylberberg2011sparse} argues that if for some batch of successive inputs, the activities of the coding neurons are uncorrelated (i.e., their computed sparse codes are uncorrelated), and all the neurons have the same average activations, the Hebbian learning rule can approximate the true stochastic gradient. Meanwhile, \cite{zylberberg2011sparse} does not provide arguments for the weight consistency between feedforward and lateral weights to be ensured by anti-Hebbian learning.
\cite{hu2014hebbian} argues that this learning framework arises from a different objective function other than~\ref{eqn:appx_dictionary_learning}. Instead, learning finds a dictionary for the following objective function
\begin{equation}
\argmin_{\mathbf{A}} \Fnormsq{\mathbf{X}^T\mathbf{X} - \mathbf{A}^T\mathbf{A}},
\end{equation}
where $\mathbf{X} \in \reals^{M\times P}$ and $\mathbf{A} \in \reals^{N\times P}$ are formed by stacking the input $\vvx$ and the sparse codes $\vva$ along the columns, respectively. This formulation is somewhat different from the dictionary learning objective function we are interested in.

The second line of work~\cite{vertechi2014unsupervised,brendel2017learning} proposes to learn the lateral weights according to the feedforward weights instead of using anti-Hebbian rules to address global weight consistency, although the learning of feedforward weight still follows Hebbian rules, giving the following update equation
\begin{equation}
\label{eqn:hebbian}
D^{({\rm new})} \leftarrow
D + \eta \left(\vv{x}\vv{a}^T - \lambda_2 D \right).
\end{equation}
It can be seen that Equation~\ref{eqn:hebbian} is not an unbiased estimate of the true stochastic gradient in Equation~\ref{eqn:appx_update_D}. 
Hence in theory the convergence of learning to an optimal solution cannot be guaranteed, and may even result in numerical instability.
Nontheless, in \cite{vertechi2014unsupervised,brendel2017learning} it was empirically shown that learning with Equation~\ref{eqn:hebbian} can progressively learn a dictionary with improved reconstruction performance if the input data is preprocessed to be whitened and centered, despite the lack of optimality guarantee.
The authors of~\cite{vertechi2014unsupervised,brendel2017learning} further propose a modified dictionary learning formulation for non-whitened input.

In this work, we propose to estimate the true stochastic gradient for dictionary learning. Therefore we do not need to make additional assumptions on the training input. 
As discussed in the main text, obtaining such estimate requires adding the feedback connections with the resulting non-trivial network dynamics. We provide extensive analysis and proofs and show that dictionary learning can be solved under this setting.

Finally, we note that the need for feedback has been repeatedly pointed out in training autoencoder networks~\cite{hinton1988learning, burbank2015mirrored}.
Autoencoder networks do not have the lateral connections as presented in the sparse coding network.
Reconstruction errors there are computed by running the network,
alternating between a forward-only and a
backward-only phase.
In contrast, we compute reconstruction errors by 
having our network evolve simultaneously with both feedforward and
feedback signals tightly coupled together.
Nevertheless, these models
do not form strong back-coupled dynamical neural networks. 
Instead, they rely on staged processing much similar to a concatenation of feedforward networks.
For our network,
the dictionary learning relies only on locations of the dynamics' trajectories at large time
which need not be close to a stable limit point.
Simple computations between these locations that corresponding to two different network 
configurations yield the necessary quantities such as reconstruction error or gradients for
minimizing a dictionary learning objective function. 

\section{Additional Numerical Experiment Results}
\subsection{Visualization of learned dictionaries}

\begin{figure*}[t]
	\centering
	\begin{tabular}{ccc}
		\includegraphics[scale=0.83]{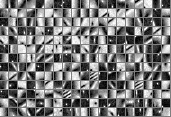} &
		\includegraphics[scale=0.26]{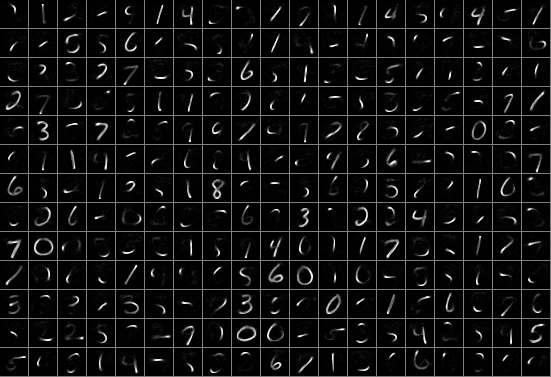} & 
		\includegraphics[scale=0.44]{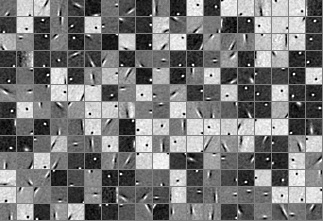} 
	\end{tabular}		
	\caption{The figure shows a random subset of the dictionaries learned in spiking networks. They show the expected patterns of edges and textures (Lena), strokes and parts of the digits (MNIST), and Gabor-like oriented filters (natural scenes), similar to those reported in prior works \cite{rubinstein2010dictionaries,ranzato2007efficient,hoyer2004non}. 
	}
	\label{fig:dictionaries}
\end{figure*}

In Section~\ref{sec:experiments}, we presented the convergence of dictionary learning by dynamical neural networks on three datasets: Lena, MNIST, and SparseNet. 
Figure~\ref{fig:dictionaries} shows the visualization of the respectively learned dictionaries.
Unsurprisingly, these are qualitatively similar to the well-known results from solving dictionary learning using canonical numerical techniques.

\subsection{Image denoising using learned dictionaries}

\begin{figure}[h]
	\centering
	\small
	\begin{tabular}{cc}
		\includegraphics[scale=0.3]{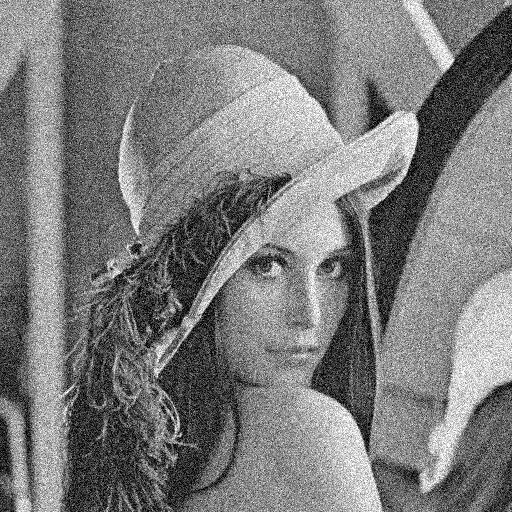} 
		&
		\includegraphics[scale=0.3]{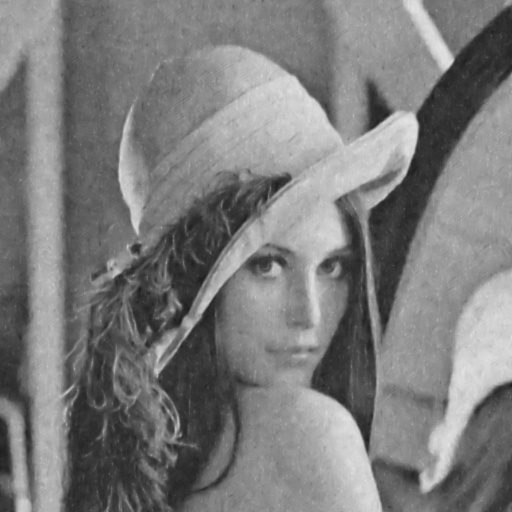} 
		\\
		(a) Noisy image (PSNR=18.69dB)
		&
		(b) Denoised image (PSNR=29.31dB)
	\end{tabular}
	\caption{Image denoising using learned dictionary.}
	\label{fig:denoising}
\end{figure}

Here we further demonstrate the applicability of the dictionary learned by our dynamical neural networks. 
We use the dictionary learned from Dataset A (the Lena image) for a denoising task using a simple procedure similar to \cite{elad2006image}:
First we extract $8 \times 8$ overlapping patches from the noisy $512 \times 512$  Lena image generated with Gaussian noise.
We then solve for the sparse coefficients of each patch in the non-negative sparse coding problem.
Using the sparse coefficients, we can reconstruct the denoised patches, and a denoised image can be obtained by properly aligning and averaging these patches. 
On average, each patch is represented by only 5.9 non-zero sparse coefficients. 
Figure \ref{fig:denoising} shows a comparison between the noisy and the denoised image.

\section{Relationships between continuous and spiking neuron model for sparse coding}

Although in this work we focus our discussions and analysis on spiking neurons, the learning strategy and mechanism can be applied to networks with continuous-valued neurons.
The close relationships between using spiking and continuous-valued neurons to solve sparse approximation problems has been discussed by \cite{shapero2014optimal,tang2016sparse}. 
Here we attempt to provide an informal discussion on the connections between the two neuron models.

Following the derivation in Section 3, the dynamics of the spiking networks can be described using the average current and spike rates.
\begin{equation}
\dot{\avgcurrvec}(t) = 
\biasvec - \avgcurrvec(t) + W\,\spikeratevec(t) + (\currvec(0)-\avgcurrvec(t))/t 
\label{eqn:spiking_dynamics}
\end{equation}
where $\avgcurrvec(t)$ and $\spikeratevec(t)$ can be related by Theorem 1 as an ``activation function''.
\begin{equation}
\spikeratevec(t) = \Thres^{-1} \max(\avgcurrvec(t), \vvzero) + \bm{\Delta}(t), \phantom{kk} \bm{\Delta}(t) \tendsto 0
\label{eqn:spiking_activation}
\end{equation}
Equation \ref{eqn:spiking_dynamics} and \ref{eqn:spiking_activation} are closedly related to the dynamics of a network of continuous-valued neuron \cite{rozell2008sparse}.
\begin{align}
\dot{\mathbf{u_c}}(t) & = \bm{\beta_c} -\mathbf{u_c}(t) + W\mathbf{a_c}(t) \\
\mathbf{a_c}(t) & = \max(\mathbf{u_c}(t),0)
\end{align}
where $\mathbf{u_c}(t)$ is the internal state variable of each neuron, $\mathbf{a_c}(t)$ is the continuous activation value of each neuron, $\bm{\beta_c}$ is the input to each neuron, and $W$ is the connection weight between neurons.
One can immediately see the similarity. Note that although such ``ReLU'' type, asymmetric activation function was not discussed in \cite{rozell2008sparse}, it was later shown in \cite{tang2016lca} that this network dynamics can solve a non-negative sparse coding problem.

\bibliographystyle{habbrv}
\bibliography{snn}

\begin{thebibliography}{10}

\bibitem{ackley1985learning}
D.~H. Ackley, G.~E. Hinton, and T.~J. Sejnowski.
\newblock A learning algorithm for {B}oltzmann machines.
\newblock {\em Cognitive science}, 9(1):147--169, 1985.

\bibitem{aharon2008sparse}
M.~Aharon and M.~Elad.
\newblock Sparse and redundant modeling of image content using an
  image-signature-dictionary.
\newblock {\em SIAM Journal on Imaging Sciences}, 1(3):228--247, 2008.

\bibitem{beck2009fast}
A.~Beck and M.~Teboulle.
\newblock A fast iterative shrinkage-thresholding algorithm for linear inverse
  problems.
\newblock {\em SIAM journal on imaging sciences}, 2(1):183--202, 2009.

\bibitem{BoydVandenberghe04}
S.~Boyd and L.~Vandenberghe.
\newblock {\em Convex Optimization}.
\newblock Cambridge University Press, Cambridge, UK, 2004.

\bibitem{brendel2017learning}
W.~Brendel, R.~Bourdoukan, P.~Vertechi, C.~K. Machens, and S.~Den{\'e}ve.
\newblock Learning to represent signals spike by spike.
\newblock {\em arXiv preprint arXiv:1703.03777}, 2017.

\bibitem{brito2016nonlinear}
C.~S.~N. Brito and W.~Gerstner.
\newblock Nonlinear {H}ebbian learning as a unifying principle in receptive
  field formation.
\newblock {\em PLoS Comput Biol}, 12(9):1--24, 2016.

\bibitem{bruckstein2008uniqueness}
A.~M. Bruckstein, M.~Elad, and M.~Zibulevsky.
\newblock On the uniqueness of nonnegative sparse solutions to underdetermined
  systems of equations.
\newblock {\em IEEE Transactions on Information Theory}, 54(11):4813--4820,
  2008.

\bibitem{burbank2015mirrored}
K.~S. Burbank.
\newblock Mirrored {STDP} implements autoencoder learning in a network of
  spiking neurons.
\newblock {\em PLoS Comput Biol}, 11(12):e1004566, 2015.

\bibitem{davies2018loihi}
M.~Davies, N.~Srinivasa, T.-H. Lin, G.~Chinya, Y.~Cao, S.~H. Choday, G.~Dimou,
  P.~Joshi, N.~Imam, S.~Jain, et~al.
\newblock Loihi: A neuromorphic manycore processor with on-chip learning.
\newblock {\em IEEE Micro}, 38(1):82--99, 2018.

\bibitem{efron2004least}
B.~Efron, T.~Hastie, I.~Johnstone, R.~Tibshirani, et~al.
\newblock Least angle regression.
\newblock {\em The Annals of statistics}, 32(2):407--499, 2004.

\bibitem{elad2006image}
M.~Elad and M.~Aharon.
\newblock Image denoising via learned dictionaries and sparse representation.
\newblock In {\em Computer Vision and Pattern Recognition, 2006 IEEE Computer
  Society Conference on}, volume~1, pages 895--900. IEEE, 2006.

\bibitem{foldiak1990forming}
P.~F{\"o}ldiak.
\newblock Forming sparse representations by local anti-{H}ebbian learning.
\newblock {\em Biological cybernetics}, 64(2):165--170, 1990.

\bibitem{freund1992unsupervised}
Y.~Freund and D.~Haussler.
\newblock Unsupervised learning of distributions on binary vectors using two
  layer networks.
\newblock In {\em Advances in neural information processing systems}, pages
  912--919, 1992.

\bibitem{hinton1988learning}
G.~E. Hinton and J.~L. McClelland.
\newblock Learning representations by recirculation.
\newblock In {\em Neural information processing systems}, pages 358--366, 1988.

\bibitem{hinton2006fast}
G.~E. Hinton, S.~Osindero, and Y.-W. Teh.
\newblock A fast learning algorithm for deep belief nets.
\newblock {\em Neural computation}, 18(7):1527--1554, 2006.

\bibitem{hopfield82}
J.~J. Hopfield.
\newblock Neural networks and physical systems with emergent collective
  computational abilities.
\newblock {\em Proc. Natl. Acad. Sci.}, 79(8):2554--2558, 1982.

\bibitem{hopfield84}
J.~J. Hopfield.
\newblock Neurons with graded response have collective computational properties
  like those of two-state neurons.
\newblock {\em Proc. Natl. Acad. Sci.}, 1:3088--3092, 1984.

\bibitem{hoyer2004non}
P.~O. Hoyer.
\newblock Non-negative matrix factorization with sparseness constraints.
\newblock {\em Journal of machine learning research}, 5(Nov):1457--1469, 2004.

\bibitem{hu2014hebbian}
T.~Hu, C.~Pehlevan, and D.~B. Chklovskii.
\newblock A hebbian/anti-hebbian network for online sparse dictionary learning
  derived from symmetric matrix factorization.
\newblock In {\em 2014 48th Asilomar Conference on Signals, Systems and
  Computers}, pages 613--619. IEEE, 2014.

\bibitem{huh2017gradient}
D.~Huh and T.~J. Sejnowski.
\newblock Gradient descent for spiking neural networks.
\newblock {\em arXiv preprint arXiv:1706.04698}, 2017.

\bibitem{kung1982systolic}
H.-T. Kung.
\newblock Why systolic architectures?
\newblock {\em IEEE computer}, 15(1):37--46, 1982.

\bibitem{lecun1998gradient}
Y.~LeCun, L.~Bottou, Y.~Bengio, and P.~Haffner.
\newblock Gradient-based learning applied to document recognition.
\newblock {\em Proceedings of the IEEE}, 86(11):2278--2324, 1998.

\bibitem{liao2016important}
Q.~Liao, J.~Z. Leibo, and T.~A. Poggio.
\newblock How important is weight symmetry in backpropagation?
\newblock In {\em AAAI}, pages 1837--1844, 2016.

\bibitem{mairal2014sparse}
J.~Mairal, F.~Bach, and J.~Ponce.
\newblock Sparse modeling for image and vision processing.
\newblock {\em Foundations and Trends{\textregistered} in Computer Graphics and
  Vision}, 8(2-3):85--283, 2014.

\bibitem{mairal2009online}
J.~Mairal, F.~Bach, J.~Ponce, and G.~Sapiro.
\newblock Online dictionary learning for sparse coding.
\newblock In {\em Proceedings of the 26th annual international conference on
  machine learning}, pages 689--696. ACM, 2009.

\bibitem{merolla2014million}
P.~A. Merolla, J.~V. Arthur, R.~Alvarez-Icaza, A.~S. Cassidy, J.~Sawada,
  F.~Akopyan, B.~L. Jackson, N.~Imam, C.~Guo, Y.~Nakamura, et~al.
\newblock A million spiking-neuron integrated circuit with a scalable
  communication network and interface.
\newblock {\em Science}, 345(6197):668--673, 2014.

\bibitem{movellan1991contrastive}
J.~R. Movellan.
\newblock Contrastive {H}ebbian learning in the continuous hopfield model.
\newblock In {\em Connectionist models: Proceedings of the 1990 summer school},
  pages 10--17, 1990.

\bibitem{olshausen1996emergence}
B.~A. Olshausen and D.~J. Field.
\newblock Emergence of simple-cell receptive field properties by learning a
  sparse code for natural images.
\newblock {\em Nature}, 381:13, 1996.

\bibitem{o1996biologically}
R.~C. O'Reilly.
\newblock Biologically plausible error-driven learning using local activation
  differences: The generalized recirculation algorithm.
\newblock {\em Neural computation}, 8(5):895--938, 1996.

\bibitem{ranzato2007efficient}
M.~Ranzato, C.~Poultney, S.~Chopra, and Y.~LeCun.
\newblock Efficient learning of sparse representations with an energy-based
  model.
\newblock In {\em Advances in neural information processing systems}, pages
  1137--1144, 2007.

\bibitem{rozell2008sparse}
C.~J. Rozell, D.~H. Johnson, R.~G. Baraniuk, and B.~A. Olshausen.
\newblock Sparse coding via thresholding and local competition in neural
  circuits.
\newblock {\em Neural computation}, 20(10):2526--2563, 2008.

\bibitem{rubinstein2010dictionaries}
R.~Rubinstein, A.~M. Bruckstein, and M.~Elad.
\newblock Dictionaries for sparse representation modeling.
\newblock {\em Proceedings of the IEEE}, 98(6):1045--1057, 2010.

\bibitem{scellier2017equilibrium}
B.~Scellier and Y.~Bengio.
\newblock Equilibrium propagation: Bridging the gap between energy-based models
  and backpropagation.
\newblock {\em Frontiers in computational neuroscience}, 11:24, 2017.

\bibitem{seung2017correlation}
H.~S. Seung and J.~Zung.
\newblock A correlation game for unsupervised learning yields computational
  interpretations of hebbian excitation, anti-hebbian inhibition, and synapse
  elimination.
\newblock {\em arXiv preprint arXiv:1704.00646}, 2017.

\bibitem{shapero2014optimal}
S.~Shapero, M.~Zhu, J.~Hasler, and C.~Rozell.
\newblock Optimal sparse approximation with integrate and fire neurons.
\newblock {\em International journal of neural systems}, 24(05):1440001, 2014.

\bibitem{tang2016lca}
P.~T.~P. Tang.
\newblock {Convergence of LCA Flows to (C)LASSO Solutions}.
\newblock {\em ArXiv e-prints}, Mar. 2016, 1603.01644.

\bibitem{tang2016sparse}
P.~T.~P. Tang, T.-H. Lin, and M.~Davies.
\newblock Sparse coding by spiking neural networks: Convergence theory and
  computational results.
\newblock {\em ArXiv e-prints}, 2017, 1705.05475.

\bibitem{vertechi2014unsupervised}
P.~Vertechi, W.~Brendel, and C.~K. Machens.
\newblock Unsupervised learning of an efficient short-term memory network.
\newblock In {\em Advances in Neural Information Processing Systems}, pages
  3653--3661, 2014.

\bibitem{whittington2017approximation}
J.~C. Whittington and R.~Bogacz.
\newblock An approximation of the error backpropagation algorithm in a
  predictive coding network with local hebbian synaptic plasticity.
\newblock {\em Neural computation}, 29(5):1229--1262, 2017.

\bibitem{xie2003equivalence}
X.~Xie and H.~S. Seung.
\newblock Equivalence of backpropagation and contrastive {H}ebbian learning in
  a layered network.
\newblock {\em Neural computation}, 15(2):441--454, 2003.

\bibitem{zylberberg2011sparse}
J.~Zylberberg, J.~T. Murphy, and M.~R. DeWeese.
\newblock A sparse coding model with synaptically local plasticity and spiking
  neurons can account for the diverse shapes of v1 simple cell receptive
  fields.
\newblock {\em PLoS Comput Biol}, 7(10):e1002250, 2011.

\end{thebibliography}

\end{document}